\documentclass[preprint, 10pt]{elsarticle}
\usepackage{lineno}
\usepackage{hyperref}
\hypersetup{hidelinks}

\usepackage{color}

\usepackage[left=2cm,right=2cm,top=2.5cm,bottom=2.5cm]{geometry}

\usepackage{booktabs}
\usepackage{graphicx}
\usepackage{subfigure}
\usepackage{epstopdf}
\usepackage{picinpar}
\usepackage{amsmath}
\usepackage{amsfonts}
\usepackage[mathscr]{euscript}
\usepackage{calligra}
\usepackage{array}
\DeclareMathAlphabet{\mathpzc}{OT1}{pzc}{m}{it}

\newcommand{\eg}{\textit{e}.\textit{g}.}

\usepackage[linesnumbered,ruled,lined]{algorithm2e}
\usepackage{amssymb}

\newdefinition{assumption}{Assumption}

\newdefinition{definition}{Definition}
\newdefinition{remark}{Remark}
\newproof{proof}{Proof}

\journal{Transportation Research Part C}

\biboptions{authoryear,compress}
\let\OLDthebibliography\thebibliography
\renewcommand\thebibliography[1]{
  \OLDthebibliography{#1}
  \setlength{\parskip}{0pt}
  \setlength{\itemsep}{0pt plus 0.3ex}
}

\begin{document}

\begin{frontmatter}

\title{Formation Control with Lane Preference for\\Connected and Automated Vehicles in Multi-lane Scenarios}

\author[thu]{Mengchi Cai}
\ead{cmc18@mails.tsinghua.edu.cn}

\author[thu]{Chaoyi Chen}
\ead{chency19@mails.tsinghua.edu.cn}

\author[thu]{Jiawei Wang}
\ead{wang-jw18@mails.tsinghua.edu.cn}

\author[thu]{Qing Xu\corref{cor1}}
\ead{qingxu@tsinghua.edu.cn}

\author[thu]{Keqiang Li\corref{cor1}}
\ead{likq@tsinghua.edu.cn}

\author[thu]{Jianqiang Wang}
\ead{wjqlws@tsinghua.edu.cn}

\author[intel]{Xiangbin Wu}
\ead{xiangbin.wu@intel.com}

\address[thu]{School of Vehicle and Mobility, Tsinghua University, Beijing, China}
\address[intel]{Intel Lab China, Beijing, China}

\cortext[cor1]{Corresponding author}

\begin{abstract}
Multi-lane roads are typical scenarios in the real-world traffic system. Vehicles usually have preference on lanes according to their routes and destinations. Few of the existing studies looks into the problem of controlling vehicles to drive on their desired lanes. This paper proposes a formation control method that considers vehicles' preference on different lanes. The bi-level formation control framework is utilized to plan collision-free motion for vehicles, where relative target assignment and path planning are performed in the upper level, and trajectory planning and tracking are performed in the lower level. The collision-free multi-vehicle path planning problem considering lane preference is decoupled into two sub problems: calculating assignment list with non-decreasing cost and  planning collision-free paths according to given assignment result. The Conflict-based Searching (CBS) method is utilized to plan collision-free paths for vehicles based on given assignment results. Case study is conducted and simulations are carried out in a three-lane road scenario. The results indicate that the proposed formation control method significantly reduces congestion and improves traffic efficiency at high traffic volumes, compared to the rule-based method.

\end{abstract}


\end{frontmatter}

%
\section{Introduction}
\label{intro}
%

Coordinated driving of Connected and Automated Vehicles (CAVs) on multi-lane roads is a critical issue for the intelligent transportation system. Existing coordination methods of multiple vehicles mainly focus on single-lane scenarios, \eg~driving as a platoon~\citep{li2017dynamical,shladover1991automated,bian2019reducing,li2020cooperative}, and intelligent intersection management where only one lane exists in each approaching direction~\citep{xu2018distributed,li2006cooperative,xu2019cooperative}. These single-lane coordination methods have revealed that coordination of multiple vehicles has great potential to improve driving safety and traffic efficiency.  However, multi-lane roads are more realistic scenarios, \eg~multi-lane ramps, lane-drop bottlenecks, and multi-lane intersections. Existing research about multi-lane coordination of CAVs mainly includes lane assignment~\citep{hall1996optimized,dao2007optimized}, merging at ramps and lane-drop bottlenecks~\citep{uno1999merging,rios2016survey}, and multi-lane formation control~\citep{marjovi2015distributed,cai2019multi}. 

Research about lane assignment aims to assign proper lanes to vehicles driving on multi-lane roads. The goal is to distribute vehicles spatially on drivable lanes, in order to fully utilize the capacity of the road and to avoid or reduce traffic congestion. One typical idea is to model the vehicle-to-lane assignment problem into a linear programming problem, by taking the percentage of vehicles to be assigned to special lanes as the programming variables~\citep{hall1996optimized,dao2007optimized,dao2008distributed}. Lane assignment methods provide optimal lane occupation suggestion for vehicles, but the lane changing control is usually performed by vehicles individually, where conflicts between vehicles may happen and traffic efficiency is limited. Multi-vehicle merging coordination methods at ramps and lane-drop bottlenecks focus on conflict resolution between vehicles driving on different lanes. Vehicles driving in these scenarios have to perform mandatory lane changing because some lanes become undrivable, \eg~the ramping lanes with finite length, and the blocked lanes at work zones~\citep{uno1999merging,zhang2019optimisation,xu2020bi}. Merging coordination methods usually set priority, calculate passing order, and select proper merging points of multiple vehicles to avoid potential collision.

\begin{figure}
\begin{center}
    \subfigure[Lane-drop bottleneck scenario]{
    \includegraphics[width=0.31\linewidth]{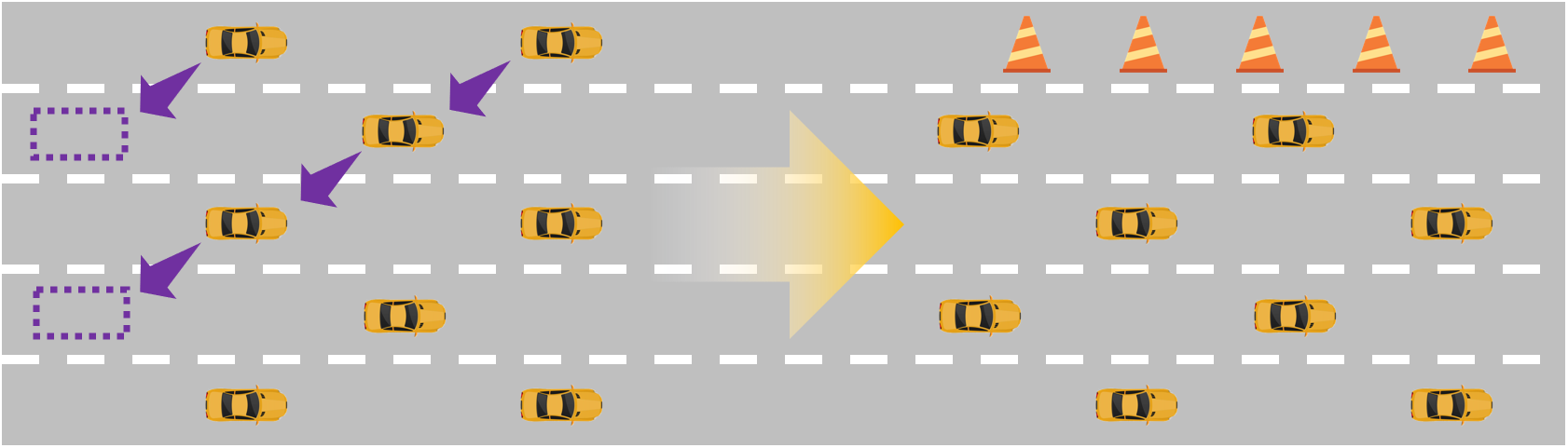}
    \label{scenario1}}
    \subfigure[Obstacle avoidance scenario]{
    \includegraphics[width=0.31\linewidth]{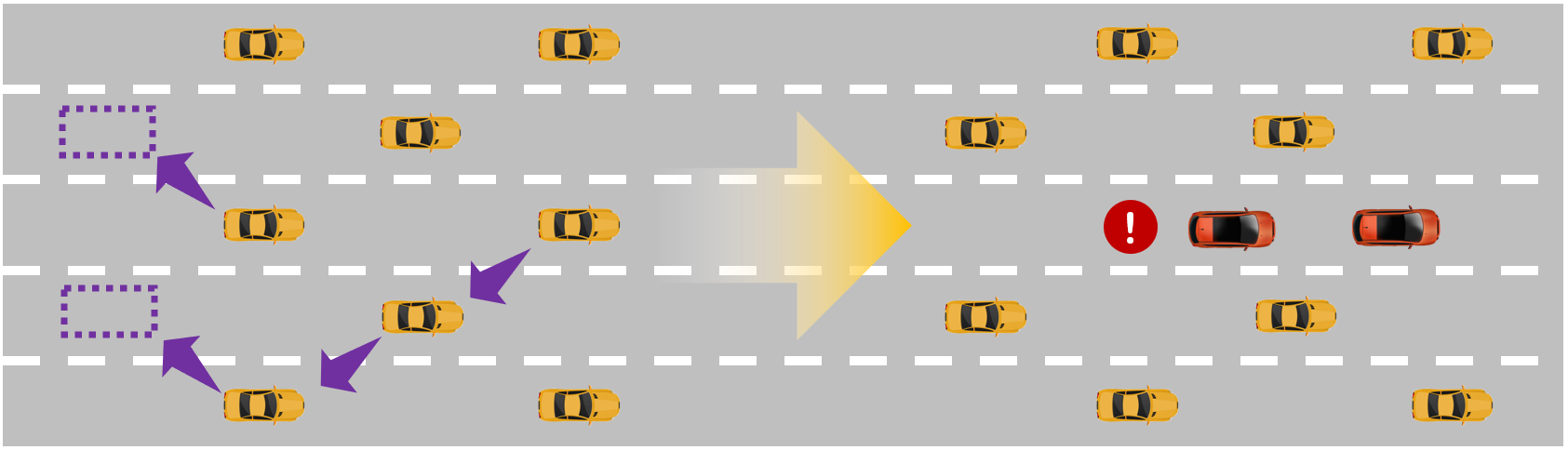}
    \label{scenario2}}
    \subfigure[Lane sorting scenario]{
    \includegraphics[width=0.31\linewidth]{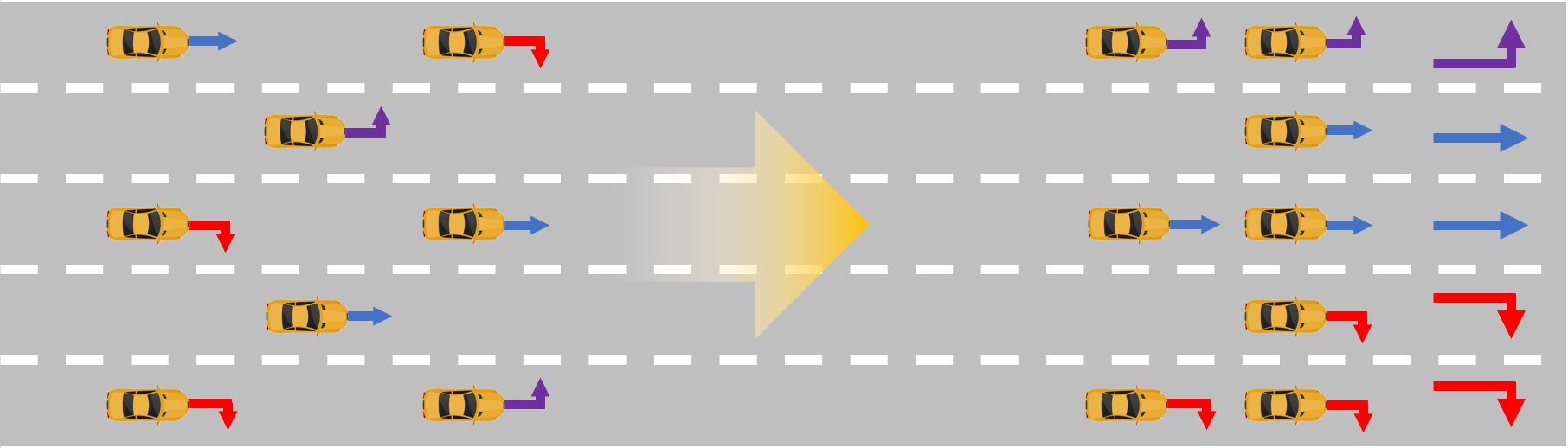}
    \label{scenario3}}
    \caption{Typical scenarios of multi-vehicle formation control. Fig.~\ref{scenario1} shows the scenario where the number of lanes changes from five to four, Fig.~\ref{scenario2} shows the scenario where one lane is blocked by two emergency vehicles, and Fig.~\ref{scenario3} shows the scenario where vehicles have preference on lanes and need to change to their target lanes.}
    \label{scenarios}
\end{center}
\end{figure}

Formation  control  (FC),  also  called  as  convoy  control, considers multiple CAVs as a group and plans the movement of vehicles from the overall perspective. The main difference between FC methods and the above on-ramp merging and lane-drop bottleneck coordination methods is that FC methods focus on the global coordination of vehicles. Vehicles drive as a group and switch the formation structure adaptively according to the changing driving environment, which enables FC methods to be applied to multiple scenarios, \eg~lane-drop bottlenecks, on-ramps, off-ramps, and lane clearance for low-speed or emergency vehicles, as shown in Fig.~\ref{scenario1} and Fig. \ref{scenario2}. The geometric structure of vehicular formation is firstly designed in~\cite{kato2002vehicle}, where vehicles follow the suggested longitudinal gaps on the multi-lane road and form interlaced formation. The slot-based approach utilizes vehicle-infrastructure  coordination to realize multi-lane driving for multiple vehicles, and is able to cover multiple scenarios, \eg~ramp merging, and formation structure switching~\citep{marinescu2010active,marinescu2012ramp}. The communication topology and consensus-based formation controller are designed in~\cite{marjovi2015distributed}, and the formation is able to avoid obstacles and switch structure at on-ramps and off-ramps. The idea of formation control and target assignment are firstly combined for robotics in~\cite{macdonald2011multi} and for vehicular formation control in~\cite{cai2019multi}. The idea is then developed to solve the multi-lane driving problem in unsignalized intersections~\citep{xu2021coordinated} and lane-drop bottlenecks~\citep{cai2021formationb}.

The aforementioned lane assignment, ramp and lane-drop bottleneck merging, and formation control methods focus on the scenarios where the drivable lanes are homogeneous. That is to say, every single vehicle can drive on every single lane, and vehicles don't have preference for special lanes. However, a more realistic scenario is that the lanes are heterogenous and have different functions. A typical example is the approaching arms of multi-lane intersections, where vehicles have different preference on special lanes and have to change to the desired lanes before certain points (\eg\ stop line of the intersection), shown in Fig.~\ref{scenario3}. In most of the research about multi-lane intersection coordination, this problem is often ignored or assumed to be solved before vehicles reaching the intersections~\citep{chen2021a,phan2020space}. In the existing few studies that focus on this problem, rule-based methods and priority-based First-In-First-Out (FIFO) methods are often considered~\citep{chouhan2020a,xu2021coordinated}. \cite{chouhan2020a} proposes a priority-based method to guide vehicles to change to their target lanes, but the lane changing time and length of road is unlimited, which makes it hard to be applied to the real world. \cite{xu2021coordinated} proposes a formation regrouping strategy to perform cooperative lane changing for vehicular formations, but the strategy is conservative and efficiency is limited.

The coordinated collision-free lane changing of multiple CAVs is similar to the collision-free coordination of intelligent agents with special targets, which is a widely studied problem in the field of robotics and artificial intelligence. Taking the desired position of vehicles after lane changing as the targets, the two problems can be transformed into each other. One of the most well-known methods that focus on multi-agent path planning is the Conflict-based Searching (CBS) method proposed in~\cite{sharon2015conflict}, where agents plan their own path to travel to the desired targets and the conflicts between agents are treated as constraints to guide the single-agent planning. The CBS method searches the global optimal solution with the minimum total cost in a built constraint tree. Some other methods are developed based on the idea of CBS to solve more complicated problems, \eg~coupled target assignment and path planning problems~\citep{ma2016optimal}, and suboptimal path planning problems~\citep{barer2014suboptimal}. Other multi-agent cooperative path planning method includes cooperative A* method~\citep{silver2005cooperative}, push and swap method~\citep{luna2011push}, and M* method~\citep{wagner2011m}.

In our previous research~\citep{cai2021formationb}, the relative formation control method is proposed to regulate and control the motion of vehicles in the Relative Coordinate System (RCS). In RCS, the relative position is spatiotemporally discretized and vehicles move on a grid map (as shown in Fig.~\ref{relative}), which is similar to the environment of multi-agent path planning. Thus, the multi-agent cooperative path planning method can be adopted for multi-vehicle cooperative lane changing.

In this paper, the bi-level motion planning framework is utilized for multi-vehicle formation control. In the upper level, vehicles' relative motion is planned in the RCS based on the CBS method. In the lower level, vehicles track the relative paths in the Geodetic Coordinate System (GCS). The focus of this paper is to calculate collision-free relative paths for vehicles moving in RCS, considering target preference (lane preference). The contributions of this paper include that:

\begin{enumerate}
\item The problem of collision-free formation control problem of multiple vehicles in GCS is modelled as the collision-free multi-agent cooperative path planning problem in RCS. The grid map in RCS is established and the relative motion of vehicles are discretized spatiotemporally. The positions of vehicles at initial and desired formation structure are projected from GCS to RCS, where collision-free relative motion of vehicles is calculated.
\item The collision-free multi-vehicle path planning problem considering lane preference is decoupled into two sub problems: calculating assignment list with non-decreasing cost and  planning collision-free paths according to given assignment result. The Conflict-based Searching (CBS) method is utilized to calculate collision-free path for vehicles in RCS. The optimality of the solution is guaranteed by the non-increasing property of assignment cost.
\item The function of the proposed method is verified in case study and the efficiency improvement is validated in simulation at different input traffic volumes. The results indicate that the rule-based method may lead to severe congestion, and the proposed formation control method is able to significantly reduce congestion and improve traffic efficiency.
\end{enumerate}

The rest of this paper is organized as follows. Section~\ref{bilevel} presents the bi-level framework for formation control. Section~\ref{model} builds the model of the formation control problem with lane preference. Section~\ref{Coupled} provides the method to decouple and solve the target assignment and collision-free path planning problem. Section~\ref{simu} carries out case study and simulations and Section~\ref{conc} gives the conclusions.

%
\section{Bi-level formation control framework}
\label{bilevel}
%
\begin{figure}
\begin{center}
    \subfigure[The bi-level framework]{
    \includegraphics[width=0.5\linewidth]{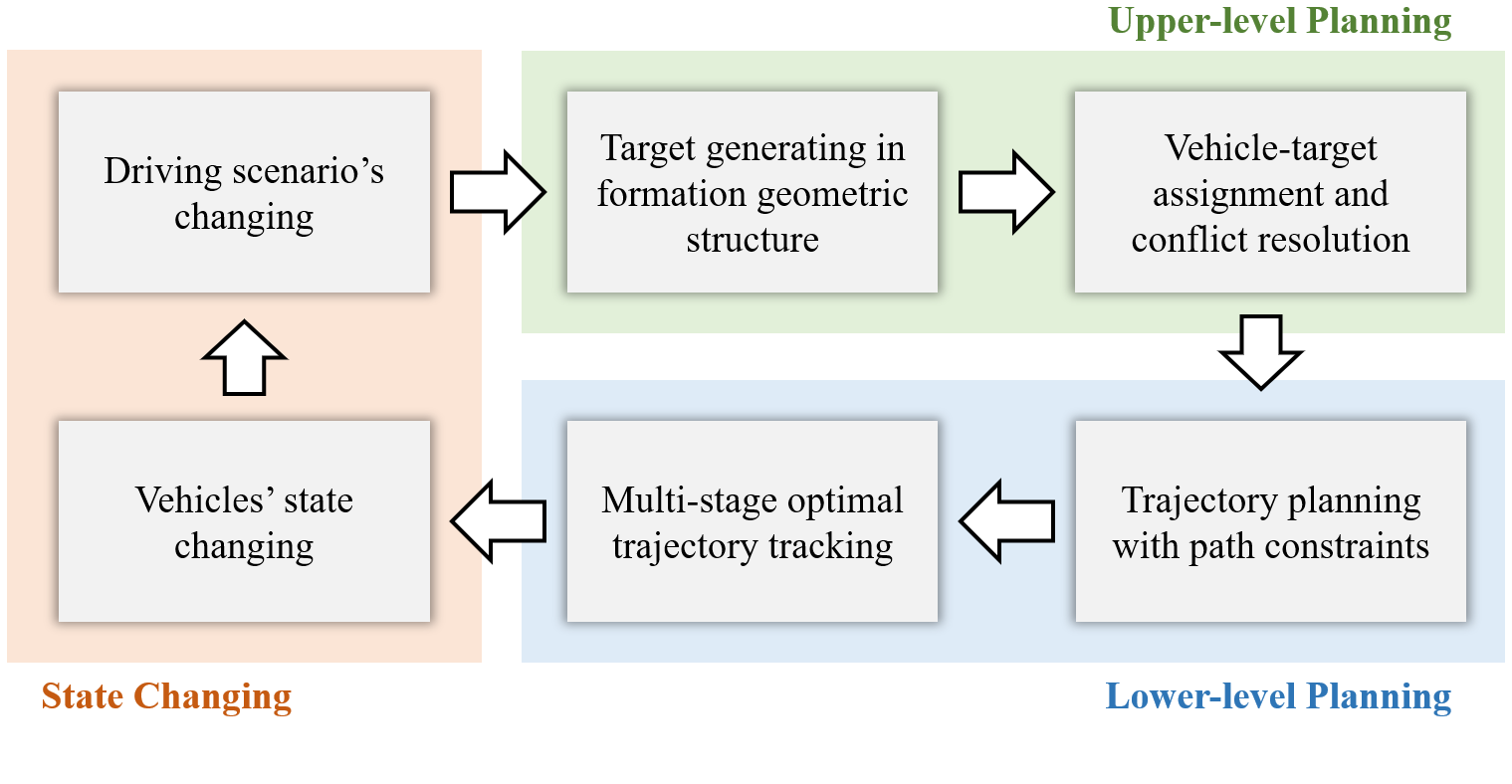}
    \label{bilevelframe}}
    \hspace{5mm}
    \subfigure[Relative and real-world state]{
    \includegraphics[width=0.41\linewidth]{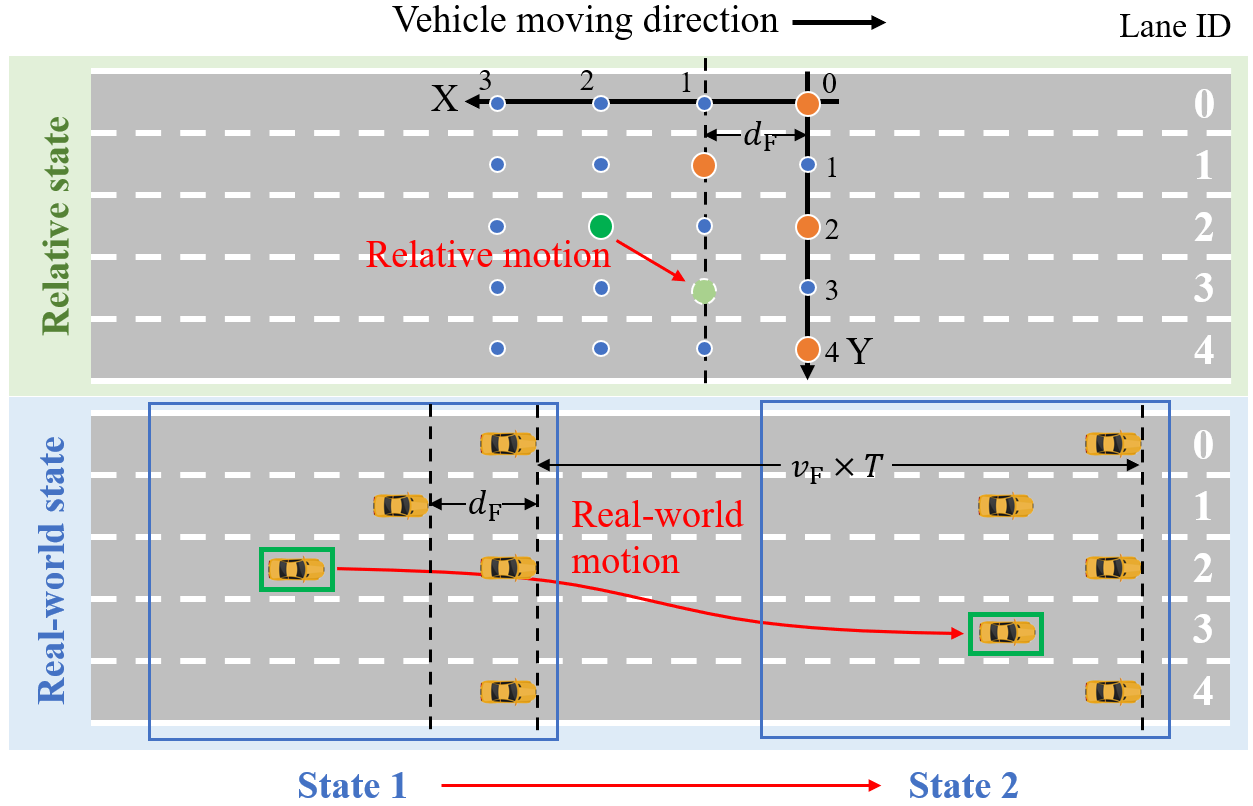}
    \label{bilevelprocess}}
    \caption{The bi-level formation control framework and formation control process. Fig.~\ref{bilevelframe} shows the working process of the upper-level and lower-level planners when the driving scenario changes. Fig.~\ref{bilevelprocess} shows the position occupation of five-vehicle formation in RCS and the relationship between relative state and real-world state.}
    \label{framework}
\end{center}
\end{figure}

The bi-level formation control framework proposed is firstly proposed in~\cite{cai2021formationb} to smoothly switch the geometric structure of vehicular formation according to the changing driving scenarios, \eg~the number of lanes changes at lane-drop bottlenecks. The framework is shown in Fig.~\ref{bilevelframe} and the state changing process in RCS and GCS is shown in Fig.~\ref{bilevelprocess}.

\subsection{Bi-level framework}
\label{bilevfram}

The bi-level formation control framework is shown in Fig.~\ref{bilevelframe}. When the driving scenario changes, the upper-level decision maker chooses the appropriate geometric structure and generates targets. Combining the information of vehicles’ state and targets’ position, the one-to-one matching between vehicles and targets is done and collision-free relative paths for vehicles to travel towards their targets are calculated. The upper-level paths are then sent to the lower-level planners. Vehicles perform trajectory planning with spatiotemporal constraints and optimal trajectory tracking in the lower level. This process changes the state of vehicles to adapt to the changing driving environment.

\begin{figure}
\begin{center}
    \subfigure[Relative coordinate system]{
    \includegraphics[width=0.42\linewidth]{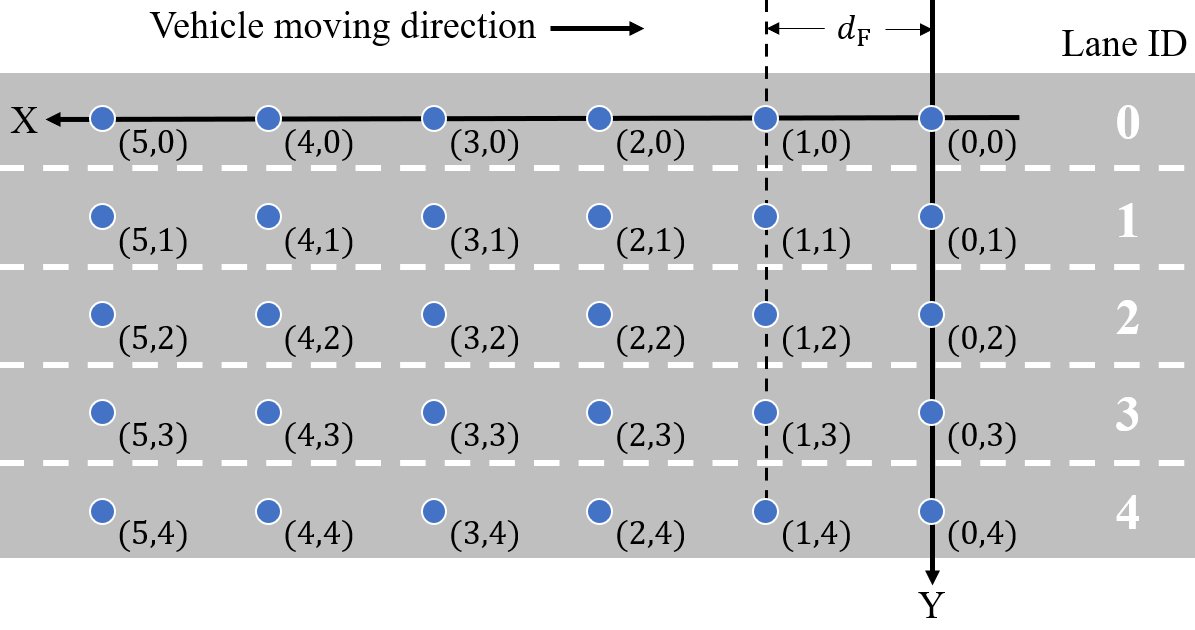}
    \label{relative}}
    \hspace{5mm}
    \subfigure[Parallel structure]{
    \includegraphics[width=0.18\linewidth]{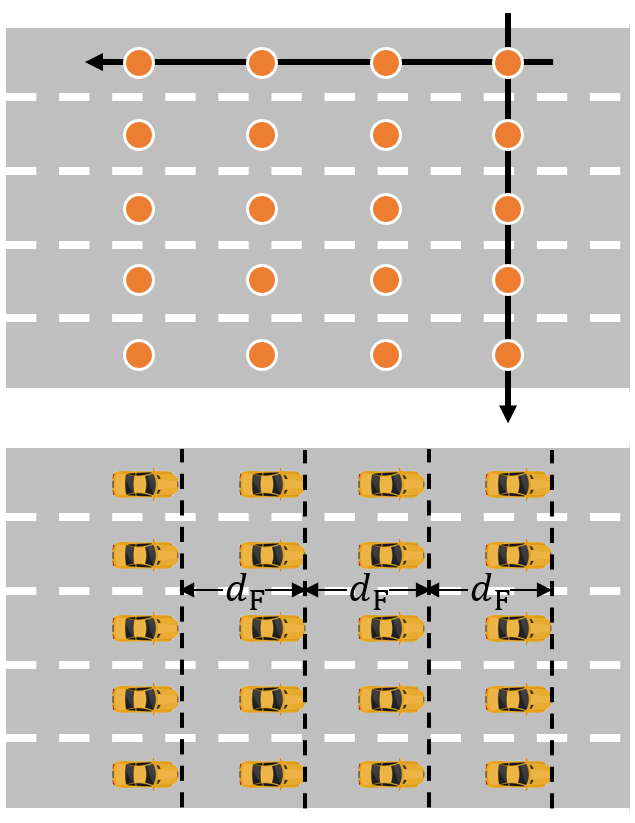}
    \label{parallel}}
    \hspace{5mm}
    \subfigure[Interlaced structure]{
    \includegraphics[width=0.18\linewidth]{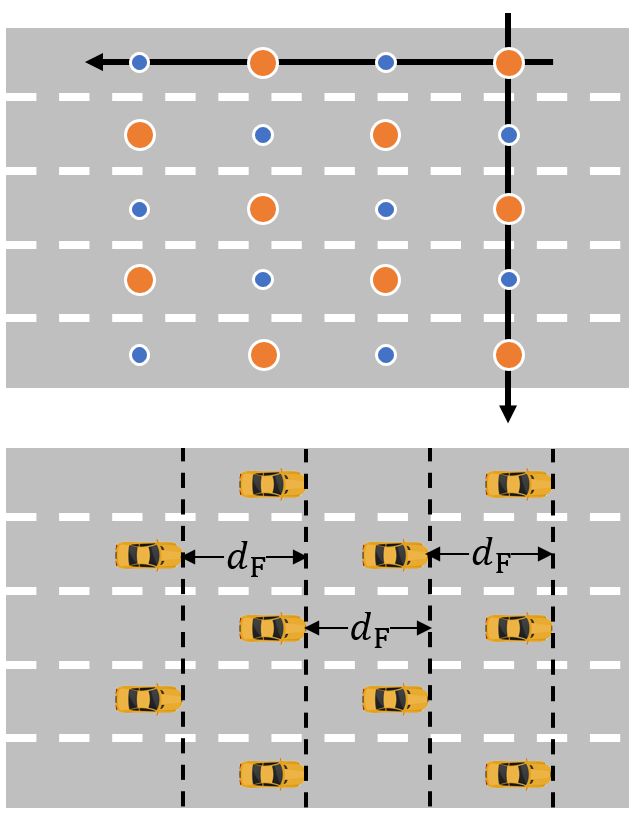}
    \label{interlaced}}
    \caption{The relative coordinate system and two typical formation structures. Fig~\ref{relative} shows the discretized relative points, and Fig~\ref{parallel} and Fig~\ref{interlaced} show the position occupation of parallel structure and interlaced structure in RCS respectively.}
    \label{relativeandstructure}
\end{center}
\end{figure}

\subsection{Upper-level relative motion planning}
\label{upper}
RCS is built to regulate vehicles' relative motion in the upper level, as shown in the upper part of Fig.~\ref{bilevelprocess}. In RCS, the perspective is changed to the moving vehicles, and the coordinate system moves with the same speed as the formation. It is assumed that the formation moves with a constant speed $v_{\text{F}}$. The relative position of vehicles in RCS is discretized, as shown in Fig.~\ref{bilevelprocess}, where blue circles represent available positions that vehicles can occupy, orange and green circles represent positions that are already occupied by vehicles. The relative position is discretized by lanes laterally, and by a constant gap $d_{\text{F}}$ longitudinally. The gap is chosen as the minimum safe following distance in a single lane. In order to describe the motion of vehicles in RCS, we firstly give the definition of path and path set. 
\begin{definition}[Path and path set]
A path is an ordered sequence of relative points in RCS that a vehicle passes through one by one. Denote the relative coordinate of vehicle $i$ at step $j$ as $(x^{\mathrm{r,v}}_{i,j},y^{\mathrm{r,v}}_{i,j})$, the path can be described as:
\begin{eqnarray}
\label{path}
P_i=\{(x^{\mathrm{r,v}}_{i,1},y^{\mathrm{r,v}}_{i,1}),(x^{\mathrm{r,v}}_{i,2},y^{\mathrm{r,v}}_{i,2}),...,(x^{\mathrm{r,v}}_{i,j},y^{\mathrm{r,v}}_{i,j})\}\ ,\ j\in \mathbb{N}^+.
\end{eqnarray}

A path set is a set of paths for a group of vehicles. The length of the paths in the same path set should be the same. A path set can be described as:
\begin{eqnarray}
\label{pathset}
\mathbb{P}=\{P_1,P_2,...,P_{N^\text{v}}\}\ ,\ N^\text{v}\in \mathbb{N}^+,
\end{eqnarray}
where $N^\text{v}$ is the number of vehicles in the group.
\end{definition}

In order to prevent collision and plan relative paths for vehicles, time is also discretized by constant time intervals $T_{\text{F}}$, and vehicles should occupy a discretized relative point at fixed time point. During a period between two time points, the relative position of a vehicle can change from one to another, or keeps unchanged. By defining the available position that a vehicle is allowed to arrive within $T_{\text{F}}$, the motion of vehicles in RCS is regulated.

\begin{definition}[Relative motion mode]
We define two modes of relative motion as follows.
\begin{enumerate}[Mode 1.]
\item The vehicles are allowed to move a longitudinal step or a lateral step within one time interval $T_{\text{F}}$. That is to say, the relative points are 4-connected. Given the relative position $(x^{\mathrm{r,v}}_{i,j},y^{\mathrm{r,v}}_{i,j})$ of vehicle $i$ at step $j$, the position $(x^{\mathrm{r,v}}_{i,j+1},y^{\mathrm{r,v}}_{i,j+1})$ at step $j+1$ is limited by:
\begin{eqnarray}
\label{4connected}
|x^{\mathrm{r,v}}_{i,j+1}-x^{\mathrm{r,v}}_{i,j}|+|y^{\mathrm{r,v}}_{i,j+1}-y^{\mathrm{r,v}}_{i,j}|\leq1.
\end{eqnarray}
\item The vehicles are allowed to move a longitudinal step, a lateral step, or a oblique step with both lateral and longitudinal movement within one time interval $T_{\text{F}}$. That is to say, the relative points are 8-connected. Given the relative position $(x^{\mathrm{r,v}}_{i,j},y^{\mathrm{r,v}}_{i,j})$ of vehicle $i$ at step $j$, the position $(x^{\mathrm{r,v}}_{i,j+1},y^{\mathrm{r,v}}_{i,j+1})$ at step $j+1$ is limited by:
\begin{eqnarray}
\label{8connected}
|x^{\mathrm{r,v}}_{i,j+1}-x^{\mathrm{r,v}}_{i,j}|\leq1,\\
|y^{\mathrm{r,v}}_{i,j+1}-y^{\mathrm{r,v}}_{i,j}|\leq1.
\end{eqnarray}
\end{enumerate}
\end{definition}

The mode of relative motion can be chosen according to the parameters and function of formation. The red arrow in the upper part of Fig.~\ref{bilevelprocess} shows an oblique move (if Relative motion mode 2 is chosen), where the vehicle moves from relative coordinate (2,2) to (1,3) to join the other four vehicles and forms an interlaced five-vehicle formation. 

As for different formation geometric structure, the position occupation in RCS is different. Parallel structure and interlaced structure are two commonly chosen formation geometric structures~\citep{kato2002vehicle,marjovi2015distributed,cai2019multi}, and the relative position occupation is shown in Fig.~\ref{relativeandstructure}. Existing research reveals that although the parallel structure has  higher  vehicle  density,  the  interlaced  structure  is  more suitable for multi-lane vehicle coordination considering lane-changing efficiency and driving safety. Thus, the interlaced structure is chosen as the standard multi-lane driving structure of vehicular formation. The choosing of geometric structure of the simulation will be further presented in Section~\ref{simu}.

\subsection{Lower-level multi-stage trajectory planning and tracking}
\label{lower}
In the lower level, the relative points in RCS are projected to real-world road points in GCS. Vehicles receive the relative paths (sequences of relative points) and calculate their real-world trajectories to path through the corresponding road points. The real-world positions of the vehicle to join the formation are marked as green rectangles in the lower part of Fig.~\ref{bilevelprocess}, and are corresponding to the green circles in the upper part. The real-world motion marked in red is the trajectory to pass through these points. This process is modelled as the optimal control problem with interior point constraints which regulate the vehicles to arrive at fixed points at fixed time. The process of the modelling and solution can be seen in~\cite{cai2021formationb}.

%
\section{Modelling of multi-vehicle formation control with lane preference}
\label{model}
%

In this section, the key is to calculate collision-free relative paths for vehicles in RCS. The initial and target positions of vehicles can be described as relative points, and the lane preference of vehicles is then transformed to the target preference. In order to calculate collision-free relative paths, the assignment problem with target preference, the conflict behavior of vehicles, and the coordinated path planning problem are modelled.

\subsection{Vehicle-target assignment with target preference}
\label{assig}
An assignment builds the one-to-one matching relationship between vehicles and targets. A match between a vehicle and a target has a corresponding cost for the vehicle to travel to the target. The assignment with the minimum total matching cost of all the vehicles is the optimal assignment. What need to be noticed is that the definition of cost is different for an assignment and a path, and is given as:
\begin{definition}[Cost]
The cost of a path represents the cost that a vehicle needs to pay when executing the path. The cost may be defined as travel time, path length, etc.  The cost of an assignment represents the estimated cost for a group of vehicles to travel to a group of targets. The cost of an assignment may be different from the total cost of paths covered by vehicles to execute the assignment, because it is common to ignore the potential collision between vehicles when calculating the cost of an assignment.
\end{definition}

In order to find the optimal assignment, the cost to assign vehicle $i$ to target $j$ should be determined at first. Among the existing research, the Euclidean Distance (ED), the square of ED, the number of lane changes, and a weighted combination of the above indexes are often taken as the assigning cost~\citep{macdonald2011multi, cai2019multi, xu2021coordinated}. In this paper, since we solve the assignment problem in RCS, the Formation Relative Distance (FRD) is chosen as the assigning cost. The FRD $d^\mathrm{r}_{i,j}$ between the vehicle $i$ whose relative coordinate is $(x^{\mathrm{r,v}}_i,y^{\mathrm{r,v}}_i)$ and the target $j$ whose relative coordinate is $(x^{\mathrm{r,t}}_j,y^{\mathrm{r,t}}_j)$ is calculated as:
\begin{eqnarray}
\label{rd}
&d^\mathrm{r}_{i,j}=l^\mathrm{o}\times N^{\mathrm{r,o}}_{i,j}+l^\mathrm{s}\times N^{\mathrm{r,s}}_{i,j},\\
&N^{\mathrm{r,o}}_{i,j}=\text{min}(|x^{\mathrm{r,v}}_i-x^{\mathrm{r,t}}_j| , |y^{\mathrm{r,v}}_i-y^{\mathrm{r,t}}_j|),\\
&N^{\mathrm{r,s}}_{i,j}=\text{max}(|x^{\mathrm{r,v}}_i-x^{\mathrm{r,t}}_j| , |y^{\mathrm{r,v}}_i-y^{\mathrm{r,t}}_j|)-\text{min}(|x^{\mathrm{r,v}}_i-x^{\mathrm{r,t}}_j| , |y^{\mathrm{r,v}}_i-y^{\mathrm{r,t}}_j|),\notag
\end{eqnarray}
where $N^{\mathrm{r,o}}_{i,j}$ and $N^{\mathrm{r,s}}_{i,j}$ represent the number of oblique and straight edges of the path respectively, and $l^\mathrm{o}$ and $l^\mathrm{s}$ represent the weight of the oblique and the straight edges. An oblique edge connects two relative points with different $X$ and $Y$ coordinates in RCS. In contrast, a straight edge connects two relative points with the same $X$ or $Y$ coordinate. In the following of this paper, the following equation holds:
\begin{eqnarray}
l^\mathrm{s}=1,\ l^\mathrm{o}=1.
\end{eqnarray}

After determining the assignment cost between each vehicle and each target, the cost matrix $\mathcal{C}$, whose element on the $i$-th row and the $j$-th column represents the cost to assign vehicle $i$ to target $j$, can be calculated as:
\begin{eqnarray}
\mathcal{C}=[c_{i,j}]\in \mathbb{R}^{N\times N},\ c_{i,j}=d^\mathrm{r}_{i,j},\ i,j\in \mathbb{N}^+.
\end{eqnarray}

Since we consider the vehicle-target assignment with target preference, the preference matrix $\mathcal{P}$ should be further defined. The element on the $i$-th row and the $j$-th column of $\mathcal{P}$ represents whether vehicle $i$ is allowed to be assigned to target $j$, and $\mathcal{P}$ is defined as:
\begin{eqnarray}
&\mathcal{P}=[p_{i,j}]\in \mathbb{R}^{N\times N}, \ 
&p_{i,j}=
\begin{cases}
1, \ \text{if vehicle $i$ is allowed to be assigned to target $j$},\notag \\
M, \ \text{otherwise},
\end{cases}
\end{eqnarray}
where $M$ is a positive number that is big enough to prevent a vehicle from being assigned to a target.

The assignment matrix $\mathcal{A}$, whose element on the $i$-th row and the $j$-th column represents whether vehicle $i$ is assigned to target $j$, is defined as:
\begin{eqnarray}
&\mathcal{A}=[a_{i,j}]\in \mathbb{R}^{N\times N}, \ 
&a_{i,j}=
\begin{cases}
1, \ \text{if vehicle $i$ is assigned to target $j$},\notag \\
0, \ \text{otherwise}.
\end{cases}
\end{eqnarray}

For simpler description of assignments, the assignment vector $\boldsymbol{a}$ is defined as:
\begin{eqnarray}
\boldsymbol{a}=[a_i]\in \mathbb{R}^{1\times N},
\end{eqnarray}
where the $i$-th element of $\boldsymbol{a}$ represents the target that is assigned to vehicle $i$. For example, if vehicle $i$ is assigned to target $j$, then $a_i$ equals to $j$.

Then, the assignment problem can be modelled as a 0-1 integer programming problem:
\begin{alignat}{2}
\min\quad & \sum_{i=1}^N\sum_{j=1}^N (c_{i,j}\times p_{i,j}\times a_{i,j}) &{}& \label{eqn - lp},\\
\mbox{s.t.}\quad
&\sum_{i=1}^N a_{i,j}=1,\notag \\
&\sum_{j=1}^N a_{i,j}=1,\notag \\
&i,j\in \mathbb{N}^+,\notag
\end{alignat}
where $N$ is the total number of vehicles, $[c_{i,j}]$ and $[p_{i,j}]$ are the given cost and preference matrices, and $[a_{i,j}]$ is the variable assignment matrix.

\subsection{Conflict model}
\label{cm}

Potential collision may happen when vehicles conduct the assignment results. In order to avoid collision, the motion of vehicles that may lead to a collision should be determined. To do this, we need to distinguish between two concepts: collision and conflict. Collision means that two vehicles collide in the real world, and the definition of conflict is given as follows.

\begin{figure}
\begin{center}
    \subfigure[Node conflict~1]{
    \includegraphics[width=0.2\linewidth]{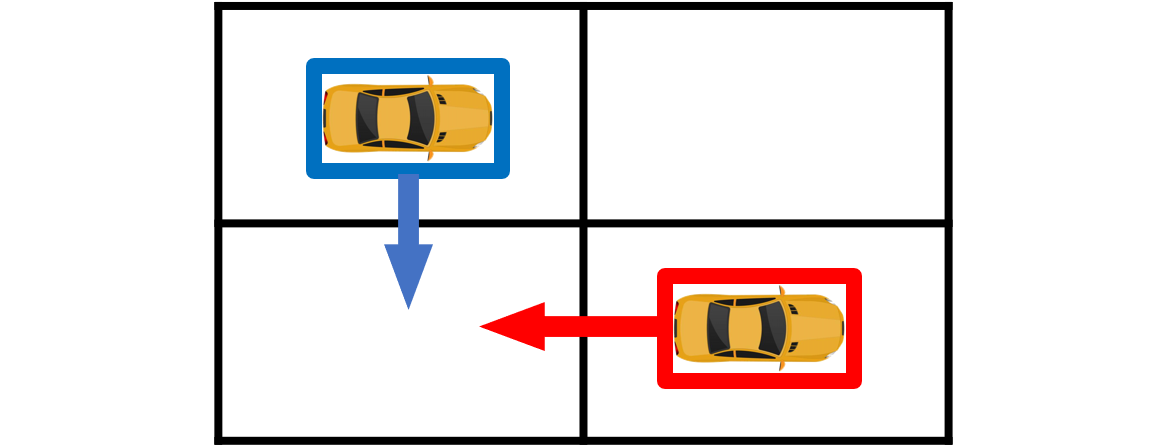}
    \label{NC1}}
    \hspace{3mm}
    \subfigure[Node conflict~2]{
    \includegraphics[width=0.2\linewidth]{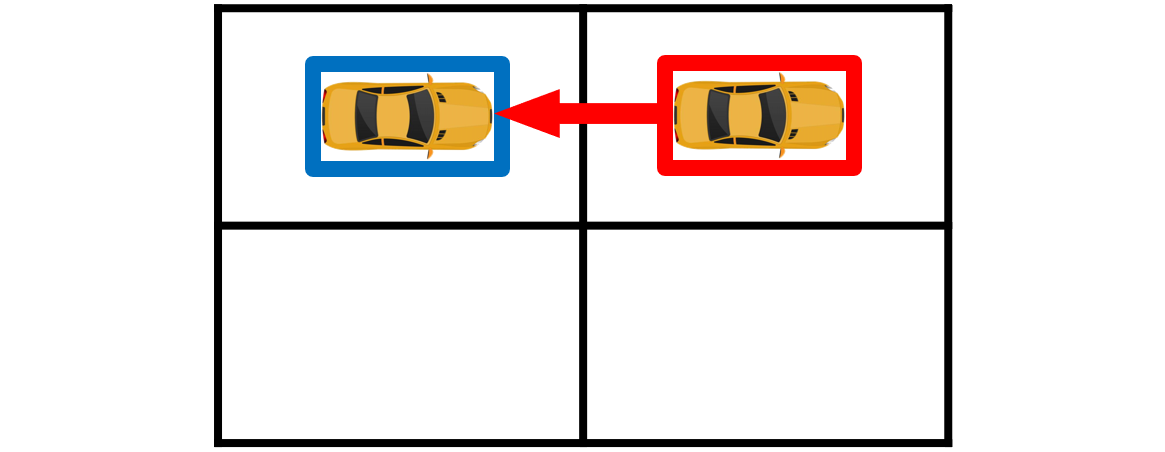}
    \label{NC2}}
    \hspace{3mm}
    \subfigure[Edge conflict~1]{
    \includegraphics[width=0.2\linewidth]{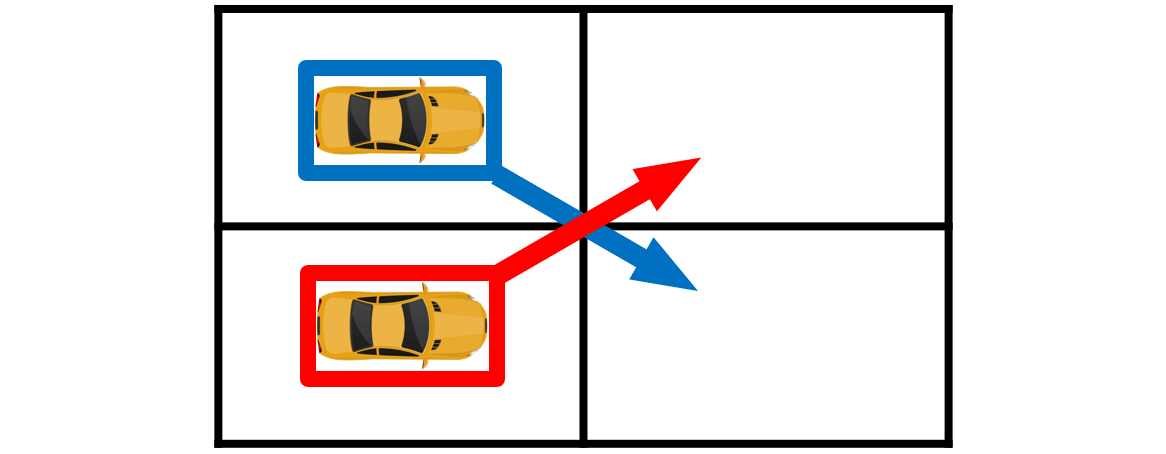}
    \label{EC1}}
    \hspace{3mm}
    \subfigure[Edge conflict~2]{
    \includegraphics[width=0.2\linewidth]{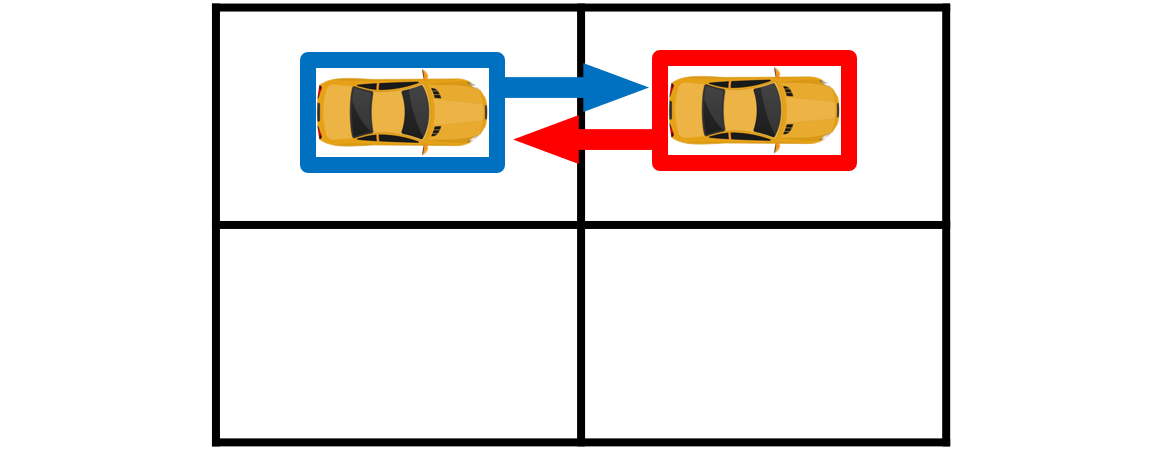}
    \label{EC2}}
    \hspace{3mm}\\
    \subfigure[Initial state of special edge conflict Type~1]{
    \includegraphics[width=0.2\linewidth]{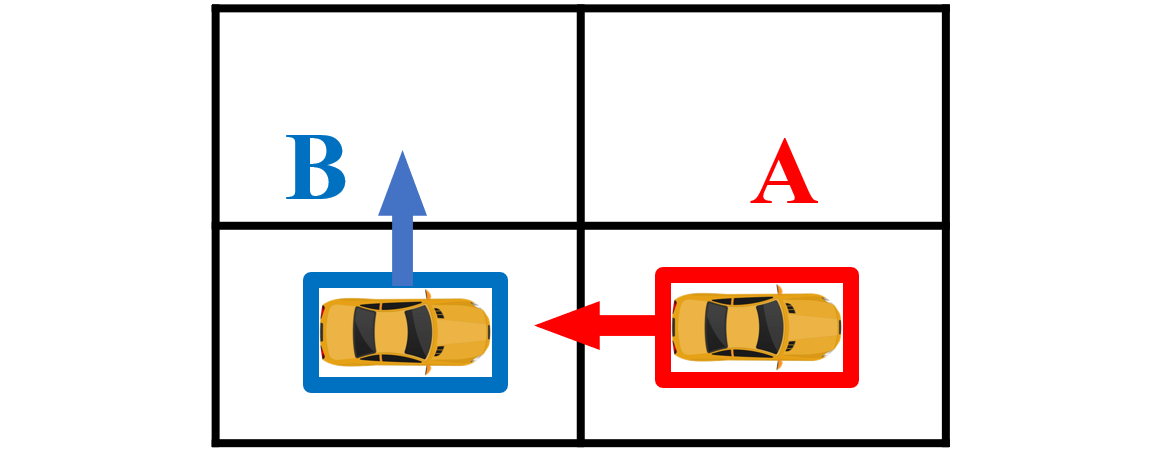}
    \label{NNC1}}
    \hspace{3mm}
    \subfigure[Intermediate state of special edge conflict Type~1]{
    \includegraphics[width=0.2\linewidth]{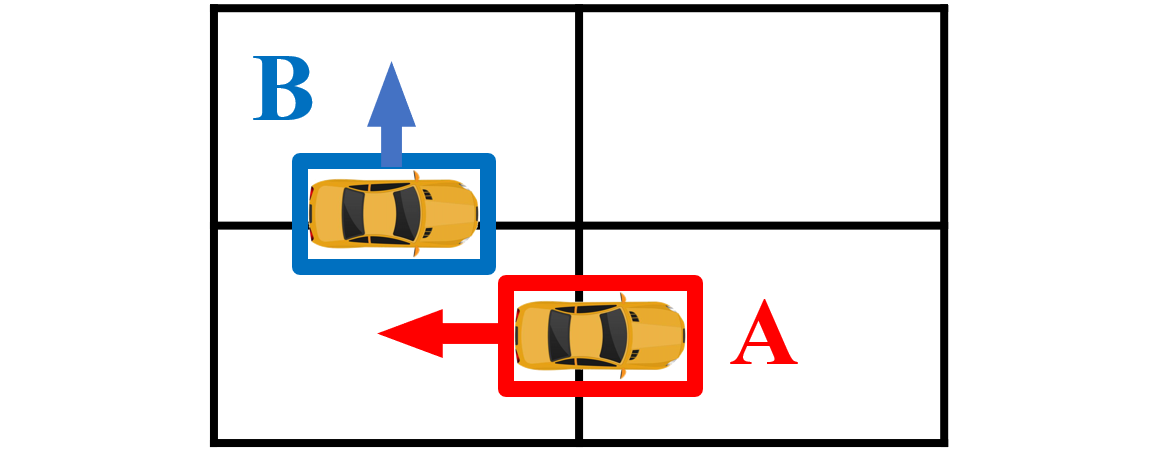}
    \label{NNC2}}
    \hspace{3mm}
    \subfigure[Initial state of special edge conflict Type~2]{
    \includegraphics[width=0.2\linewidth]{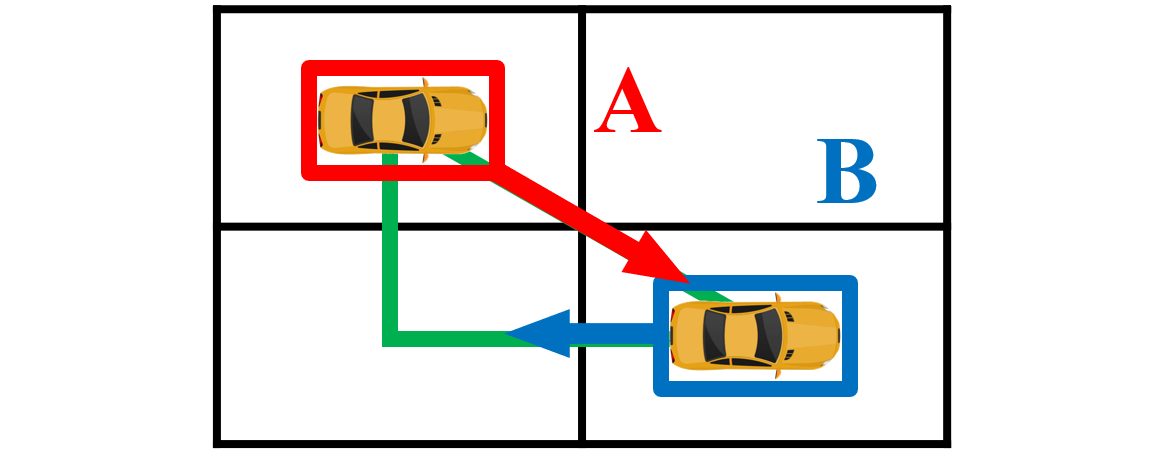}
    \label{ECS11}}
    \hspace{3mm}
    \subfigure[Intermediate state of special edge conflict Type~2]{
    \includegraphics[width=0.2\linewidth]{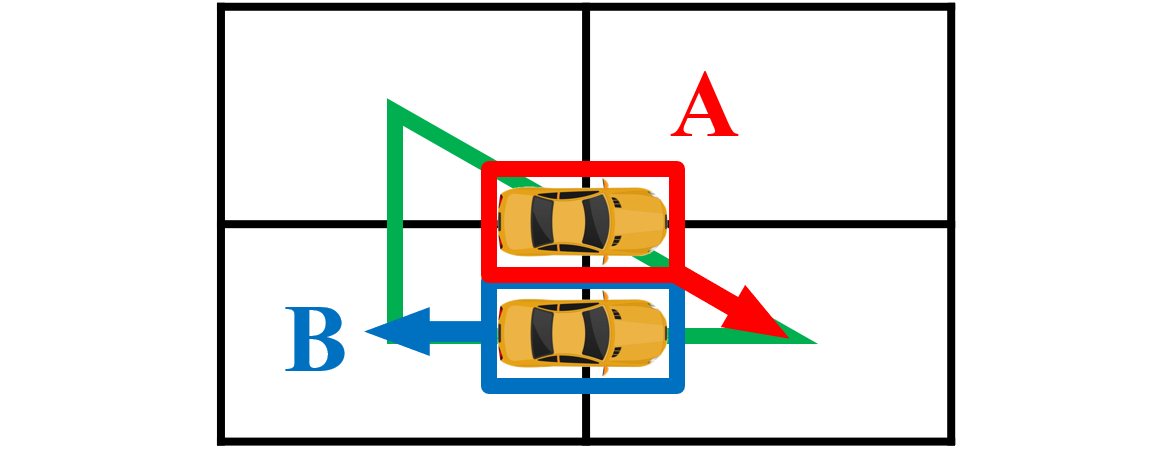}
    \label{ECS12}}
    \hspace{3mm}\\
    \subfigure[Initial state of special edge conflict Type~3]{
    \includegraphics[width=0.2\linewidth]{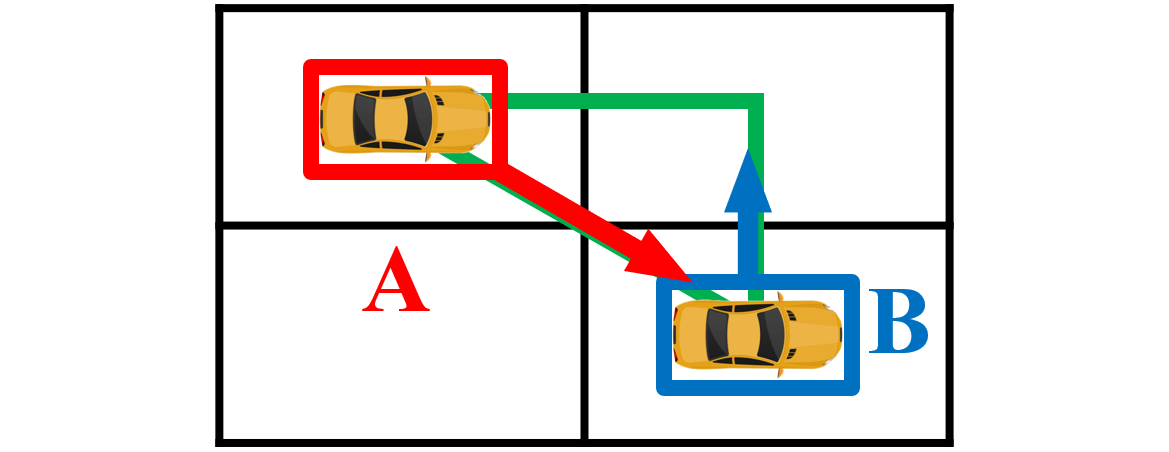}
    \label{ECS21}}
    \hspace{3mm}
    \subfigure[Intermediate state of special edge conflict Type~3]{
    \includegraphics[width=0.2\linewidth]{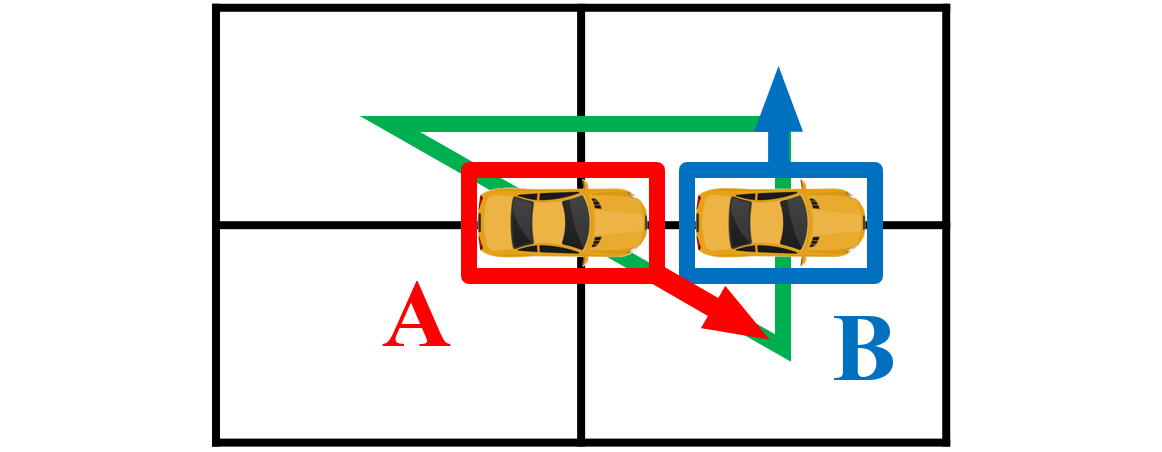}
    \label{ECS22}}
    \hspace{3mm}
    \subfigure[Initial state of special edge conflict Type~4]{
    \includegraphics[width=0.2\linewidth]{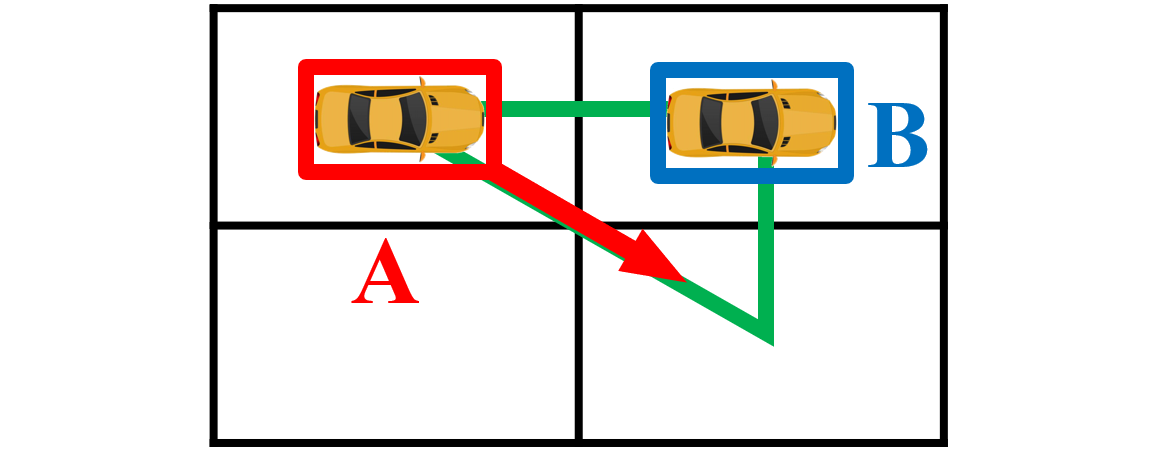}
    \label{ENC1}}
    \hspace{3mm}
    \subfigure[Intermediate state of special edge conflict Type~4]{
    \includegraphics[width=0.2\linewidth]{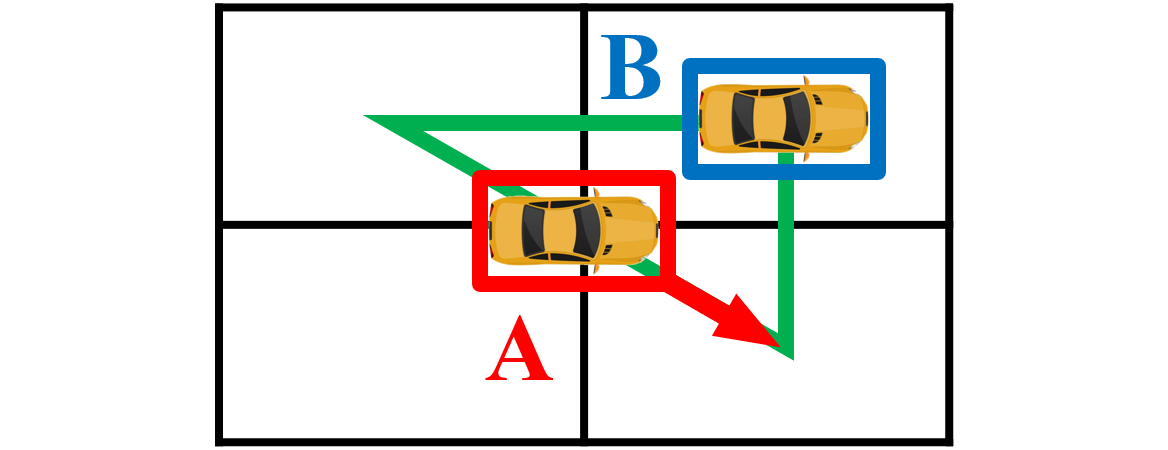}
    \label{ENC2}}
    \caption{Motion conflicts that may lead to collision between vehicles moving in RCS.}
    \label{conflicts}
\end{center}
\end{figure}

\begin{definition}[Conflict]
A conflict represents that the motion of two or more vehicles may lead to collision. It is defined according to the position and time that the potential collision may happen. Generally, there are two types of conflicts: the node conflict and the edge conflict. 

If two vehicles arrive at the same relative point at the same time, a collision may happen and the two vehicles are in node conflict. A node conflict may happen because of two reasons: one vehicle moves to the point that another vehicle is already occupied and maintained, and two vehicles move to the same relative points from other two different points, as shown in Fig.~\ref{NC1} and Fig.~\ref{NC2}. 

If the paths of two vehicles during the same time interval $T_{\text{F}}$ cross or overlap, a collision may happen along the edge and the two vehicles are in edge conflict, as shown in Fig.~\ref{EC1} and Fig.~\ref{EC2}. 

Since the size of vehicles can't be ignored in the real world, some behavior of vehicles may still cause collision, even though the relative trajectories of vehicles don't intersect. Thus, some special edge conflicts are further defined. In this paper, the following four types of special edge conflicts are considered:
\begin{enumerate}[Type 1.]
\item If vehicle A moves to the current relative position of vehicle B and vehicle B leaves the position and moves to another point in the same time interval, vehicle A and vehicle B are in the first type of special edge conflicts, as shown in Fig.~\ref{NNC1} and Fig.~\ref{NNC2}.
\item If vehicle A moves an oblique step and vehicle B moves a longitudinal step and the initial and target positions of these two vehicles form a triangle, vehicle A and vehicle B are in the second type of special edge conflicts, as shown in Fig.~\ref{ECS11} and Fig.~\ref{ECS12}.
\item If vehicle A moves an oblique step and vehicle B moves a lateral step and the initial and target positions of these two vehicles form a triangle, vehicle A and vehicle B are in the third type of special edge conflicts, as shown in Fig.~\ref{ECS21} and Fig.~\ref{ECS22}.
\item If vehicle A moves an oblique step and vehicle B maintains its current position and the initial and target positions of these two vehicles form a triangle, vehicle A and vehicle B are in the fourth type of special edge conflicts, as shown in Fig.~\ref{ENC1} and Fig.~\ref{ENC2}.
\end{enumerate}

\end{definition}

It should be noticed that whether the defined four types of special edge conflicts will cause collision is determined according to the driving environment. If the lane is narrow and the following gap (longitudinal discrete interval) is short, all the four special edge constraints may cause collision and should be avoided. However, if the lane is wide enough, type 2 may not cause collision, and if the following gap is long enough, type 1, type 3, and type 4 may not cause collision. Moreover, only type 1 should be considered for relative motion mode 1, and all the four types should be considered for relative motion mode 2.

\subsection{Relative path map}
\label{rpp}
Relative path map is a clear way to show the moving process and examine whether a conflict exists~\citep{cai2021formationb}. Put the ID and the path of vehicle $i$ on the $i$-th row, we can get the relative path map for the $N^\mathrm{v}$ vehicles, as is shown in Table \ref{rpm}. The elements in the same column represent the relative positions of vehicles at the same fixed time point. Table \ref{rpm} shows the steps that each vehicle will take to travel to its assigned target, which makes it clear to find and resolve the potential collision between vehicles during the formation switching process.

\begin{table*}[htbp]
\begin{center}
\caption{Relative path map}
\label{rpm}
\begin{tabular}{cccccccccccc}
\toprule
Vehicle ID              &step 1  &step 2 & ...   &step $j$& ...&step $N^\mathrm{s}$\\
\midrule
1     &   $(x^{\mathrm{r,v}}_{1,1},y^{\mathrm{r,v}}_{1,1})$    & $(x^{\mathrm{r,v}}_{1,2},y^{\mathrm{r,v}}_{1,2})$& ... & $(x^{\mathrm{r,v}}_{1,j},y^{\mathrm{r,v}}_{1,j})$& ...& $(x^{\mathrm{r,v}}_{1,N^\mathrm{s}},y^{\mathrm{r,v}}_{1,N^\mathrm{s}})$\\ 
2     &   $(x^{\mathrm{r,v}}_{2,1},y^{\mathrm{r,v}}_{2,1})$   & $(x^{\mathrm{r,v}}_{2,2},y^{\mathrm{r,v}}_{2,2})$& ... & $(x^{\mathrm{r,v}}_{2,j},y^{\mathrm{r,v}}_{2,j})$& ...& $(x^{\mathrm{r,v}}_{2,N^\mathrm{s}},y^{\mathrm{r,v}}_{2,N^\mathrm{s}})$\\ 
...                 & ...                          & ...        &   ...      & ...  & ...      & ... \\ 
$i$        &   $(x^{\mathrm{r,v}}_{i,1},y^{\mathrm{r,v}}_{i,1})$     & $(x^{\mathrm{r,v}}_{i,2},y^{\mathrm{r,v}}_{i,2})$& ... & $(x^{\mathrm{r,v}}_{i,j},y^{\mathrm{r,v}}_{i,j})$& ...& $(x^{\mathrm{r,v}}_{i,N^\mathrm{s}},y^{\mathrm{r,v}}_{i,N^\mathrm{s}})$\\ 
...              & ...                    & ...      &   ...       & ...    & ...   & ... \\ 
$N^\mathrm{v}$ &$(x^{\mathrm{r,v}}_{N^\mathrm{v},1},y^{\mathrm{r,v}}_{N^\mathrm{v},1})$&$(x^{\mathrm{r,v}}_{N^\mathrm{v},2},y^{\mathrm{r,v}}_{N^\mathrm{v},2})$& ...& $(x^{\mathrm{r,v}}_{N^\mathrm{v},j},y^{\mathrm{r,v}}_{N^\mathrm{v},j})$& ...& $(x^{\mathrm{r,v}}_{N^\mathrm{v},N^\mathrm{s}},y^{\mathrm{r,v}}_{N^\mathrm{v},N^\mathrm{s}})$\\
\bottomrule  
\end{tabular}
\end{center}
\end{table*}

%
\section{Decoupled target assignment and path planning}
\label{Coupled}
%

In this section, we design the framework to decouple target assignment and path planning problems, and calculate the optimal path solution. We firstly give the decoupled framework, and then provide candidate assignment calculating and collision-free path planning methods.

\subsection{Framework of decoupled target assignment and path planning}
\label{ctapp}

\begin{figure}
\begin{center}
    \includegraphics[width=0.5\linewidth]{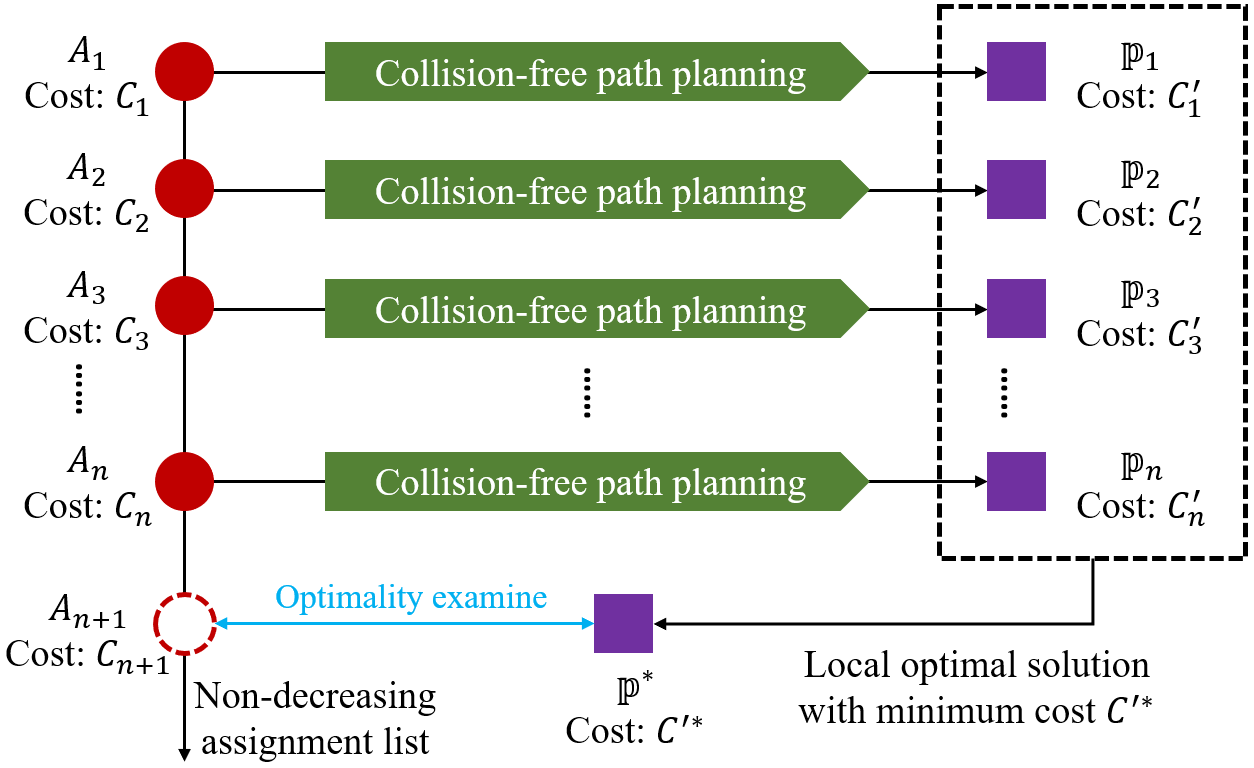}
    \caption{The framework of solving the coupled optimal target assignment and path planning problem. The red circles represent assignment candidates with a non-increasing order of total cost. In the green parts the conflict-free coordinated path planning is performed, and the sets of paths are got, shown as purple rectangles.}
    \label{frame}
\end{center}
\end{figure}

The vehicle-to-target assignment is modelled in Section~\ref{model}. Since different assignment leads to different relative paths which have different cost, the target assignment and path planning problem should be considered together. The framework of decoupled target assignment and path planning is shown in Fig.~\ref{frame}. The red circles show a list of assignments with non-decreasing cost, the greed parts show the process of collision-free path planning, and the purple rectangles represent the path solutions. The assigning cost of the optimal assignment $A_1$ is $C_1$, and the corresponding path set is $\mathbb{P}_1$. Since the assigning cost between a vehicle and a target is calculated without considering other vehicles, the cost of the collision-free path, noted as $C_1'$, is possibly bigger and always not smaller than $C_1$. Thus, it is not sure whether $\mathbb{P}_1$ is the optimal path solution with the smallest cost, so a new assignment $A_2$, which is the next optimal assignment but $A_1$, is calculated and the new path set $\mathbb{P}_2$ is planned. After calculating $n$ assignments and path sets, the local optimal path solution $\mathbb{P}^*$ with current minimum cost $C'^*$ is got. If the assigning cost $C_{n+1}$ of next optimal assignment $A_{n+1}$ is bigger or equal to $C'^*$, it is not possible to find an $A_{n+1}$-based path set with a smaller cost than $C'^*$, and $\mathbb{P}^*$ is the optimal path solution. Otherwise, the process steps forward and get a new local optimal path solution. The process is given in Algorithm~\ref{tappa}. CBSA in the fifth and seventeenth line is short for conflict-based searching algorithm, which is given in Algorithm~\ref{cbsa}.

\begin{algorithm}[tb]
\label{tappa}
\caption{The Target Assignment and Path Planning Algorithm (TAPPA)}  
\LinesNumbered  
\KwIn {Initial position of vehicles.\newline
Target position of vehicles.
}
\KwOut{Optimal path set $\mathbb{P}^*$ with the minimum cost.
}  
\textbf{Initialization} Set $k\leftarrow1$, $f\leftarrow0$, $\mathbb{A}\leftarrow\emptyset$, $\mathbb{P}^*\leftarrow\emptyset$, $C'^*\leftarrow+\infty$.\\
Calculate the optimal assignment $\mathcal{A}_k$, and get the cost $C_{k}$ of $\mathcal{A}_{k}$.\\
$\mathbb{A}\leftarrow \mathbb{A}\cup\{\mathcal{A}_{k}\}$.\\
\While{The next optimal assignment $\mathcal{A}_{k+1}$ but $\mathbb{A}$ exists,}{
Apply CBSA to $\mathcal{A}_k$ to get the path set $\mathbb{P}_k$, and get the cost $C_{k}'$ of $\mathbb{P}_k$.\\
\If{$C_{k}'<C'^*$}{
Set $C'^*\leftarrow C_{k}'$, $\mathbb{P}^*\leftarrow\mathbb{P}_k$.
}
Calculate the next optimal assignment $\mathcal{A}_{k+1}$ but $\mathbb{A}$, and get the cost $C_{k+1}$ of $\mathcal{A}_{k+1}$.\\
\If{$C'^*\leq C_{k+1}$}{
Set $f\leftarrow1$.\\
\textbf{break}
}
$k\leftarrow k+1$.\\
}
\If{$f=0$}{
Apply CBSA to $\mathcal{A}_k$ to get the path set $\mathbb{P}_k$, and get the cost $C_{k}'$ of $\mathbb{P}_k$.\\
\If{$C_{k}'<C'^*$}{
Set $C'^*\leftarrow C_{k}'$, $\mathbb{P}^*\leftarrow\mathbb{P}_k$.
}
}
\end{algorithm}

\subsection{Candidate assignment calculating}
\label{candi}
The assignment list with non-decreasing cost is calculated one by one, starting from the global optimal assignment. Many classical algorithms are designed to solve this kind of problem, and Hungarian Algorithm is one of the most commonly used one~\citep{kuhn1955hungarian}. When calculating the next optimal assignment, we can remove the global optimal assignment from the solution space and calculate the local optimal assignment in the rest of the solution space using the same algorithm as calculating the global optimal solution. There have been numerous research about calculating the optimal or next optimal assignment solution~\citep{murthy1968algorithm,chegireddy1987algorithms}, and we omit the details in this paper.

\subsection{Collision-free path planning method}
\label{cppm}
\begin{figure}
\begin{center}
    \includegraphics[width=0.8\linewidth]{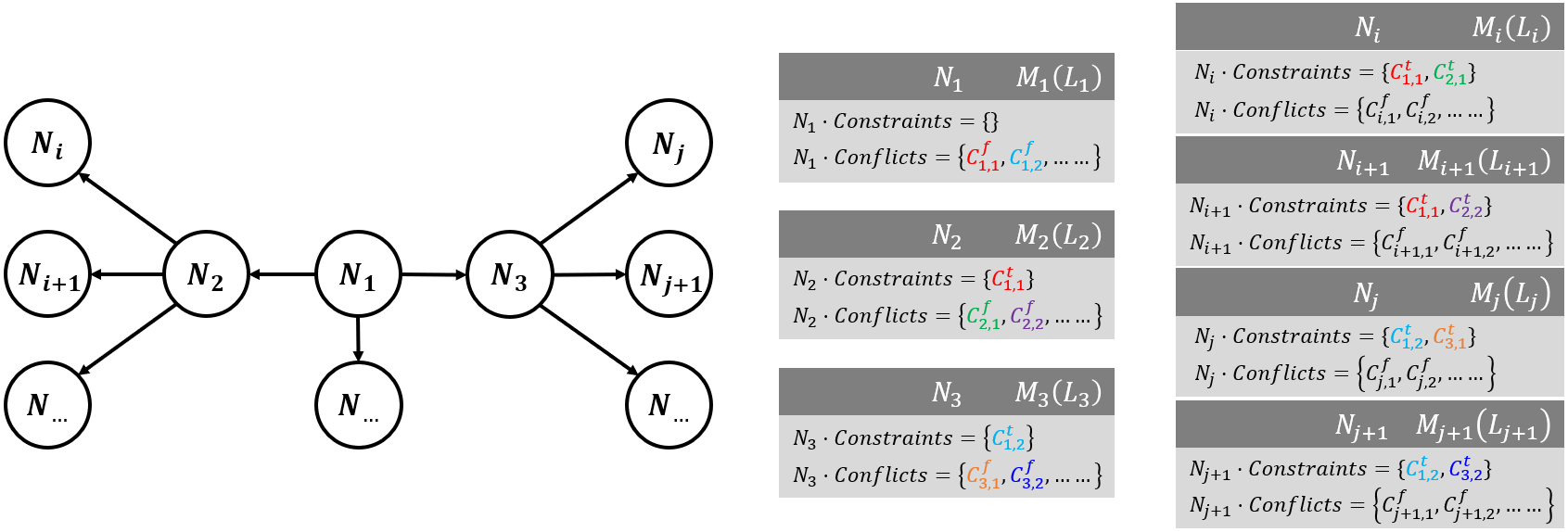}
    \caption{The constraint tree built during the process of CBS method.}
    \label{cTree}
\end{center}
\end{figure}

In this paper, the Conflict-based Searching (CBS) method~\citep{sharon2015conflict} is utilized to calculate collision-free paths for vehicles in RCS. The CBS method searches the optimal paths with the minimum total cost in a conflict tree, as shown in the left part of Fig.~\ref{cTree}. Each node in the tree has four attributes, including constraints, conflicts, path set, and cost, as shown in the right part of Fig.~\ref{cTree}. 

The definition of conflict has been given in Section~\ref{cm}. When a collision may happen for vehicle $i$ at step $j$, the conflict is denoted as $C^f_{i,j}$. A node conflict is described as $C^f_{i,j}=\{\text{N}|(x,y), step, id\}$, where $(x,y)$ is the coordinated of the conflict node, $step$ is the time point when the conflict happens, and $id$ is the ID of the vehicle. An edge conflict is described as $C^f_{i,j}=\{\text{E}|(x_1,y_1), (x_2,y_2), step, id\}$, where $(x_1,y_1)$ and $(x_2,y_2)$ are the coordinate of the starting point and the end point of the conflict edge respectively, and $step$ is the time point when the vehicle starts to travel along the conflict edge. We further give the definition of constraint.

\begin{definition}[Constraint]
A constraint is defined in the searching tree to prevent certain vehicle to perform certain motion. A constraint is generated according to a conflict. If the conflict is a node conflict, the constraint prevents the vehicle to travel to the node at the given time point. If the conflict is an edge conflict, the constraint prevents the vehicle to travel along the edge during the given time interval. The constraint generated according to conflict $C^f_{i,j}$ is denoted as $C^t_{i,j}$. A constraint can be described as the same equation as the corresponding conflict.
\end{definition}

\begin{algorithm}[tb]
\label{cbsa}
\caption{The Conflict-based Searching Algorithm (CBSA)}
\LinesNumbered
\KwIn {Initial position of vehicles.\newline
Target position of vehicles.\newline
A given assignment.
}
\KwOut{Optimal path set $\mathbb{P}^*$ with the minimum cost.
}
\textbf{Initialization} Set $\mathbb{T}\leftarrow\emptyset$.\\
Set ${N_1}.{Constraints}\leftarrow\emptyset$.\\
Apply SPPA to ${N_1}$ to calculate path set ${M_1}$ with cost ${L_1}$.\\
Calculate ${N_1}.{Conflicts}$ based on ${M_1}$.\\
Set $\mathbb{T}\leftarrow\mathbb{T}\cup\{N_1\}$.\\
\While{$\mathbb{T}$ is not empty,}{
Find ${N_i}$ in $\mathbb{T}$ with the minimum $L_i$.\\
Set $\mathbb{T}\leftarrow\mathbb{T}\setminus\{N_i\}$.\\
\If{${N_i}.Conflicts=\emptyset,$}{
Set $\mathbb{P}^*\leftarrow M_i$.\\
\textbf{break}
}
\ForEach{element $C^f_{i,j}$ in ${N_i}.{Conflicts}$}{
Add a child node ${N_{c}}$ to ${N_i}$.\\
Calculate constraint $C^t_{i,j}$ based on conflict $C^f_{i,j}$.\\
Set ${N_{c}}.{Constraints}\leftarrow{N_i}.{Constraints}\cup C^t_{i,j}$.\\
Apply SPPA to ${N_c}$ to calculate path set ${M_c}$ with cost ${L_c}$.\\
Calculate ${N_c}.{Conflicts}$ based on ${M_c}$.\\
Set $\mathbb{T}\leftarrow\mathbb{T}\cup\{N_i\}$.
}
}
\end{algorithm}  

For each node in the searching tree, individual path planning is performed for every vehicle, considering the constraints. The path planning is conducted without considering other vehicles, and may cause conflicts, which is then added as the attributes of the node. The edge between nodes represents the generating order of different nodes. Two nodes are called as parent node and child node if there is a directed edge connecting them and starting from the parent node to the child node. All the constraints of a parent node should also be added to its child node, and one of the conflicts of the parent node will be added as a new constraint to its child node. That is to say, the number of child nodes of a node is equal to the number of the parent node's conflicts. The cost of a node is the total cost of the path set calculated according to the constraints. Note that not every cost is solid, because there may be conflicts and the corresponding path set calculated for the node will cause collision. If there is no conflict for a node, this node is said to be a feasible solution, but not always an optimal solution. If there is no conflict for a node and its cost is not smaller than any other feasible nodes, this node is said to be an optimal solution with the minimum cost.

The tree starts from the root node $N_1$, which doesn't have constraints since it has no parent node. By applying single-vehicle path planning method to the initial and target position of vehicles, a path set, a conflict set, and a cost is got. The single-vehicle path planning algorithm utilized in this paper is the A* algorithm~\citep{hart1968formal}. Child nodes are generated to the root node and the tree iterates the process until an optimal solution is found. The process of CBS method is given in Algorithm~\ref{cbsa}. SPPA in the third line is short for single-vehicle path planning algorithm.

The CBS method is well known and widely used in the field of multi-robot path planning. The properties of CBS method including completeness and time complexity have been proved and explained in numerous papers~\citep{sharon2013increasing,sharon2015conflict}, and we omit them in this paper.

%
\section{Simulations and Results}
\label{simu}
%

In this section, simulations are carried out and the results are analyzed. Firstly, a case is presented to show the process of the proposed method in detail. Then, the scenario and parameter settings are presented and the reference rule-based method is introduced. The performance of the proposed and reference methods is compared and the reasons are explained.

\subsection{Case study}
\label{case}
\begin{figure}
\begin{center}
    \subfigure[The position of vehicles]{
    \includegraphics[width=0.2\linewidth]{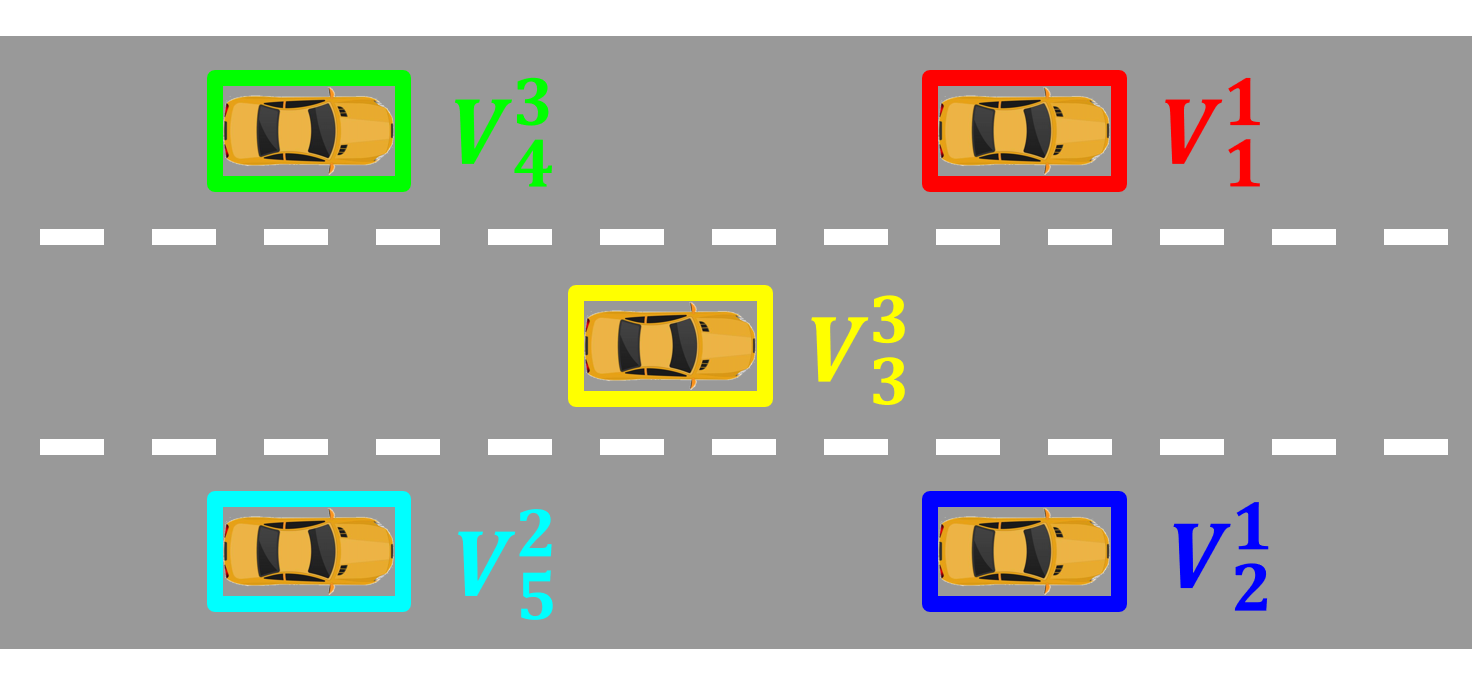}
    \label{fe1v}}
    \hspace{5mm}
    \subfigure[The position of targets]{
    \includegraphics[width=0.2\linewidth]{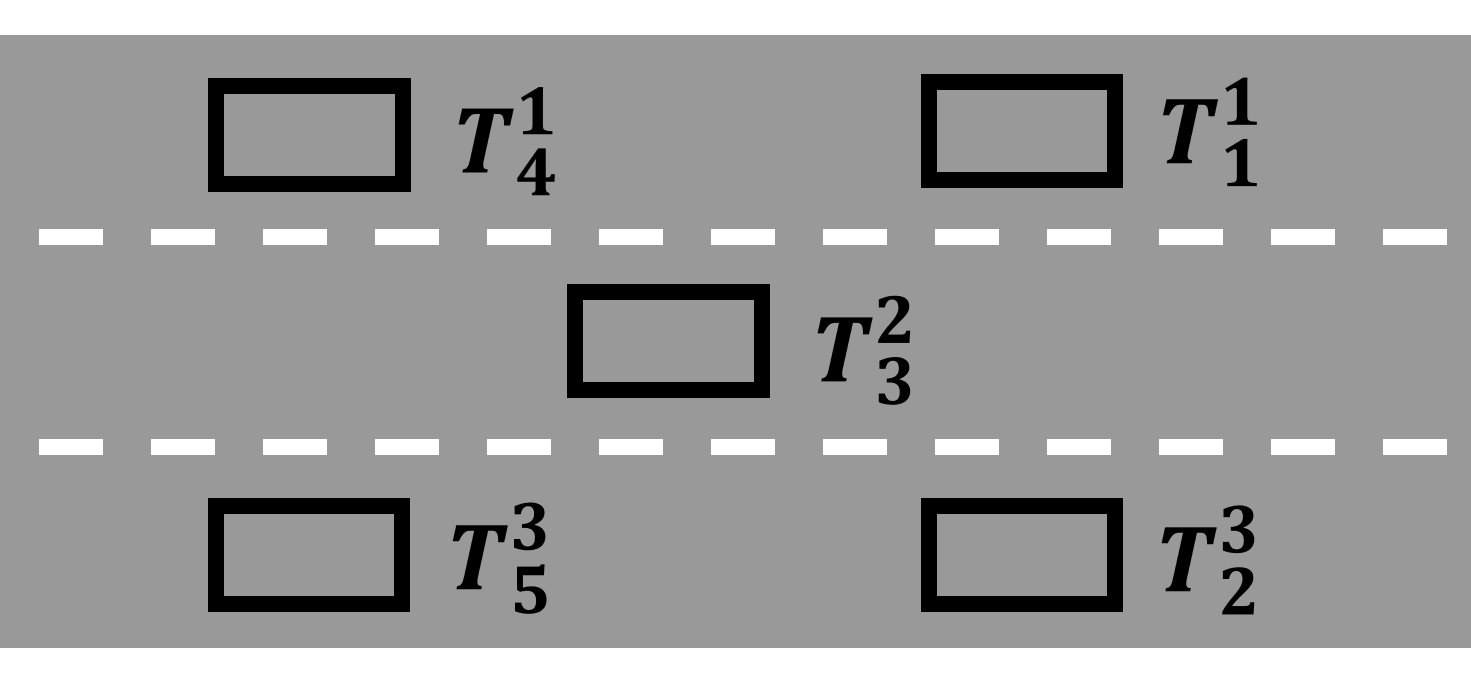}
    \label{fe1t}}
    \caption{The initial and target position of vehicles in case study.}
    \label{e1}
\end{center}
\end{figure}

In this part, a five-vehicle formation switching example is given to describe the building process of the conflict tree in detail. The formation contains five vehicles in an interlaced structure, as shown in Fig.~\ref{e1}. The vehicles and targets are represented as $V_i^j$ and $T_i^j$ respectively, and $i$ represents the ID and $j$ represent the preference of vehicles on targets. Vehicles and targets with the same $j$ are allowed to be assigned to each other.

\begin{figure}
\begin{center}
    \subfigure[Solving process]{
    \includegraphics[width=0.7\linewidth]{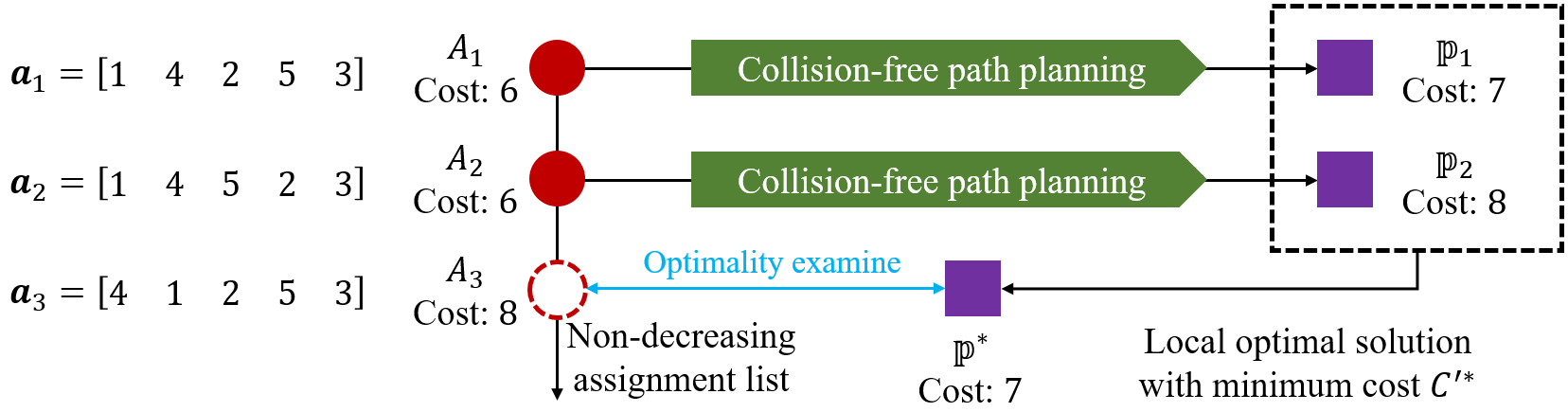}
    \label{framee1}}\\
    \subfigure[The three-layer tree]{
    \includegraphics[width=0.9\linewidth]{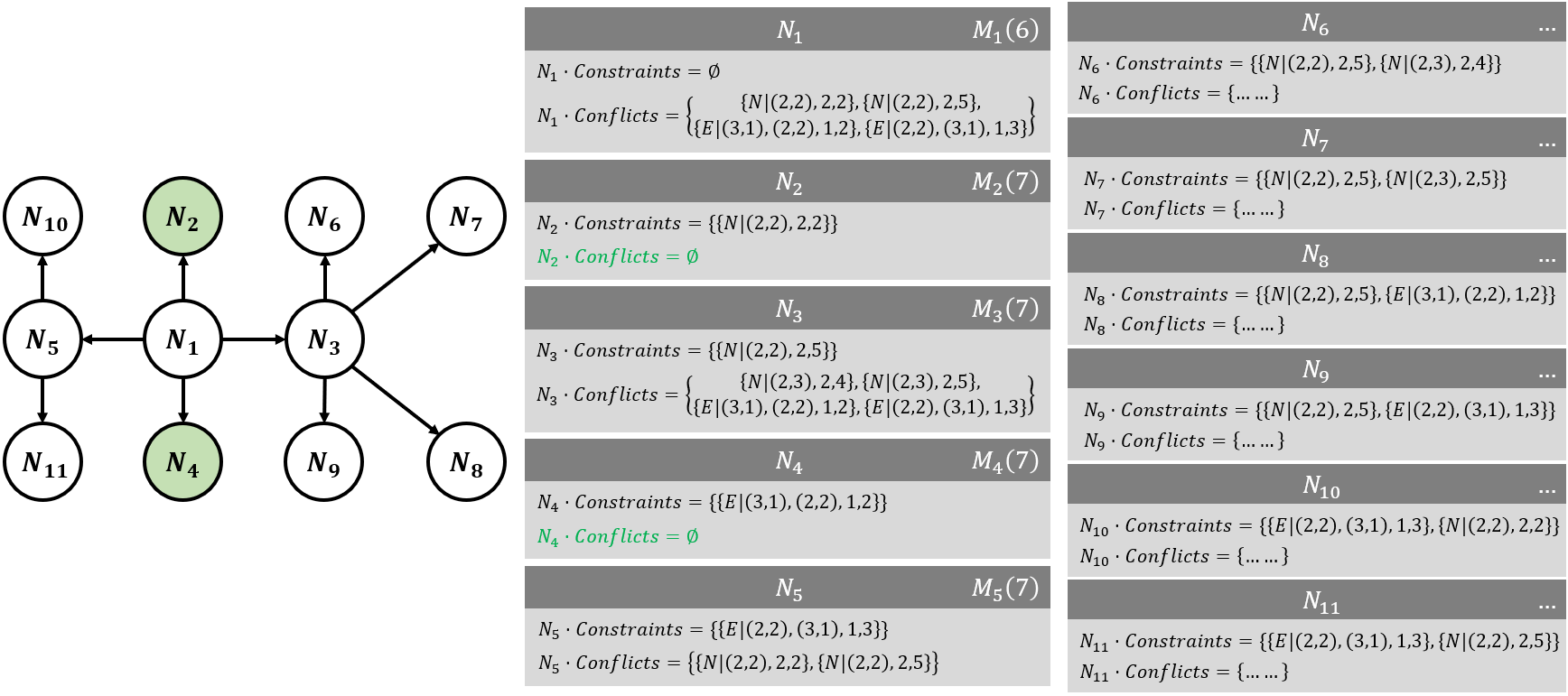}
    \label{cTree12}}
    \subfigure[Relative path maps]{
    \includegraphics[width=0.75\linewidth]{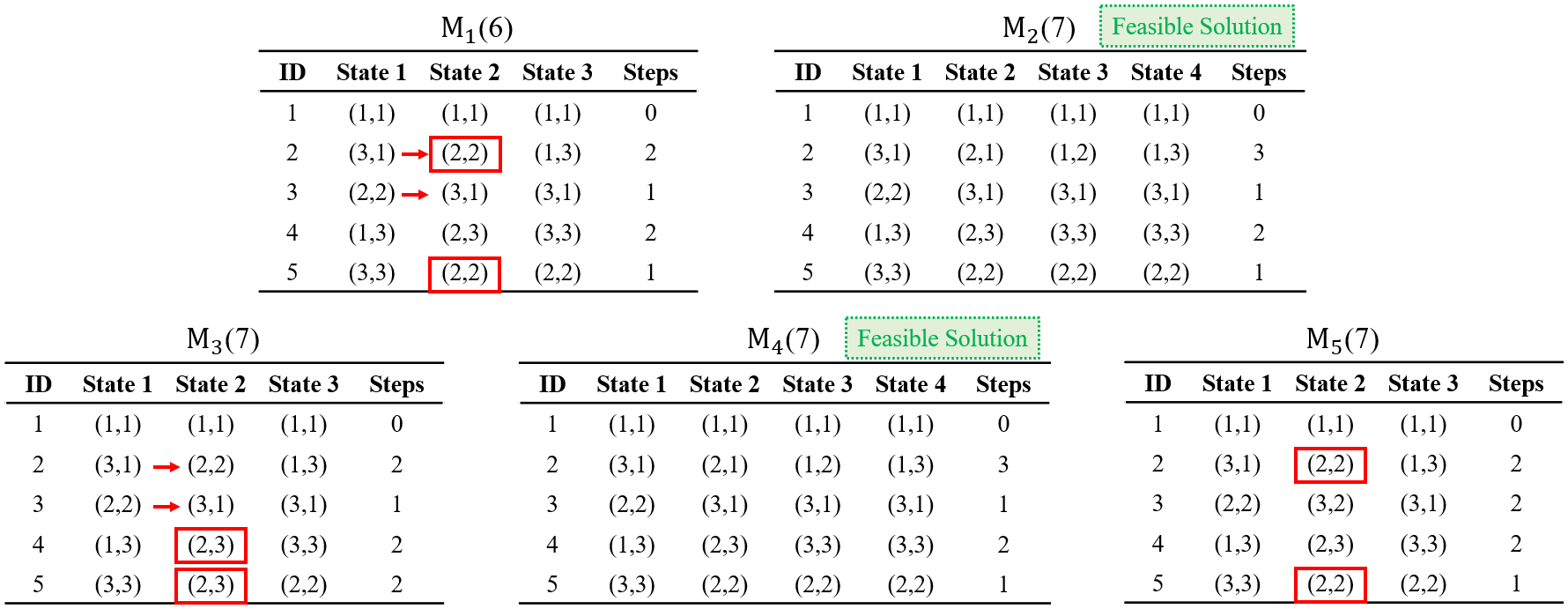}
    \label{cMapse1}}
    \caption{The process of CBS method in case study.}
    \label{cTree1}
\end{center}
\end{figure}

The optimal assignment calculated for this case is $A_1$ with assignment vector $\boldsymbol{a}_1=[1,4,2,5,3]$ and cost $C_1=6$, and the collision-free path planning is conducted. The decoupled target assignment and path planning framework of this case is shown in Fig.~\ref{framee1}. A root node $N_1$ is generated by applying A* algorithm for every vehicle to plan a path to their assigned targets, without considering other vehicles. The relative path map is built to show the formation switching process, as shown in Fig.~\ref{cMapse1}. A node conflict is detected for vehicle 2 and vehicle 5 at the first step between state 1 and state 2, and an edge conflict is detected for vehicle 2 and vehicle 3 at the same step. Then the four conflicts are added for $N_1$ and accordingly, four child nodes are generated by adding the conflicts of $N_1$ as the constraints of the new nodes. The four child nodes of $N_1$, noted as $N_2$, $N_3$, $N_4$, and $N_5$, are then applied with single-vehicle path planning algorithm, and the result shows that $N_2$ and $N_4$ are feasible solution because no conflicts are detected, and new conflicts are detected for $N_3$ and $N_5$, and child nodes are accordingly generated, as shown in Fig.\ref{cTree12}. Since the cost of child node are not smaller than that of parent node, $N_2$ and $N_4$ with cost 7 are both optimal solution of assignment $A_1$. Note that Algorithm~\ref{cbsa} will terminate once a solution is found, which means that the process will end when either $N_2$ or $N_4$ is tested with no conflicts, and the testing of other nodes is skipped. Here, in the case study, we provide the complete process of testing all the four child nodes of $N_1$ to show the process of the CBS method more clearly.

\begin{figure}
\begin{center}
    \includegraphics[width=0.7\linewidth]{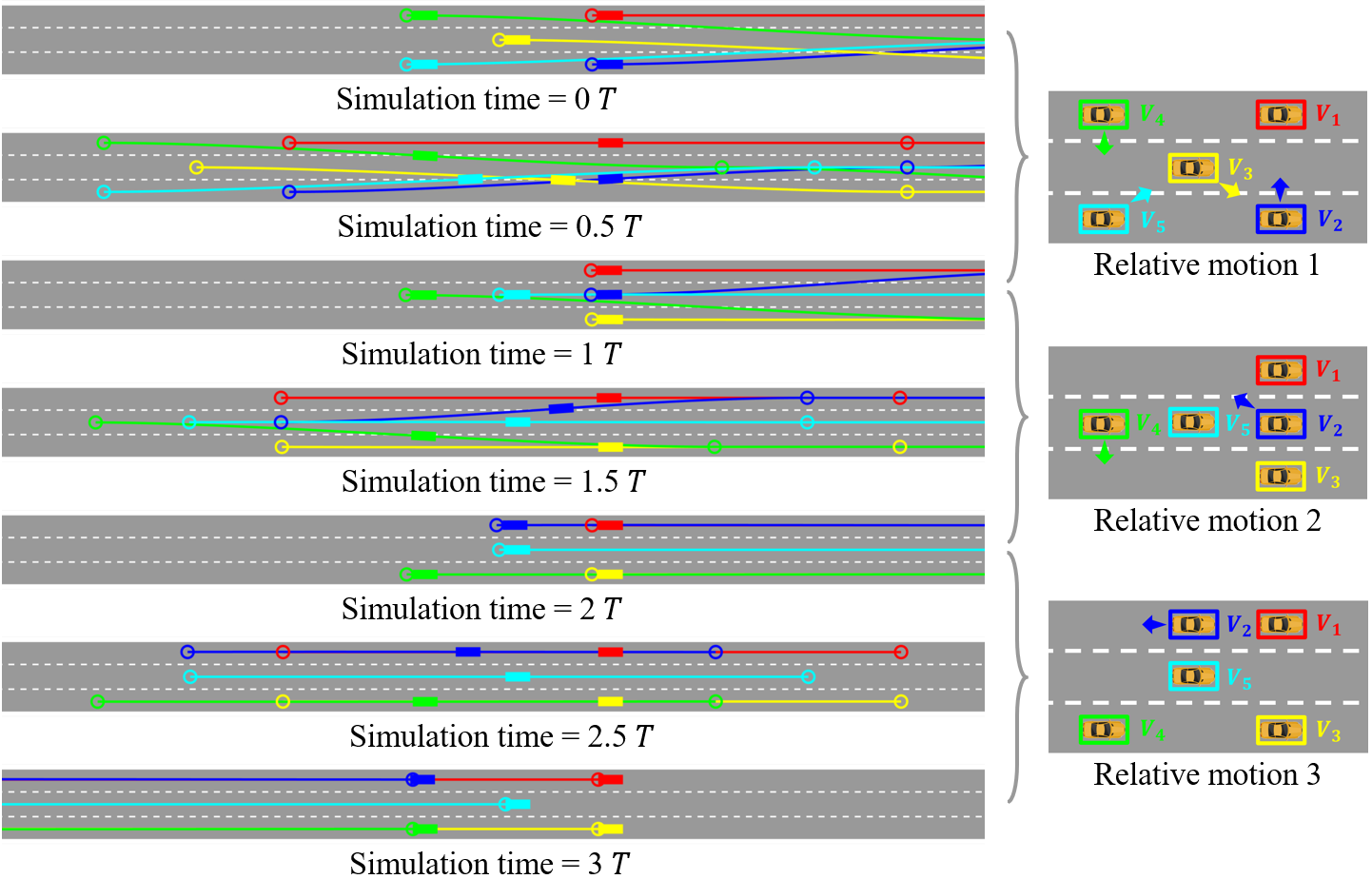}
    \caption{The trajectories and relative motion of formation switching process. The full version video is available online at the address: {\color{blue}https://github.com/cmc623/Formation-control-with-lane-preference}}
    \label{casee1}
\end{center}
\end{figure}

Since the cost of the path set $\mathbb{P}_1$ is 7 and is bigger than the cost of $A_1$, we don't know whether $\mathbb{P}_1$ is the global optimal solution, and the next optimal assignment $A_2$ is calculated and the cost is $C_2=6$. The method of calculating optimal and next optimal assignment has been given in Section~\ref{candi}. It indicates that $A_2$ has the potential to generate a better solution than $\mathbb{P}_1$ because the cost of $A_2$ is smaller than that of $\mathbb{P}_1$. After path planning, the optimal solution for $A_2$ is $\mathbb{P}_2$ with cost 8, so $\mathbb{P}_1$ remains as the current local optimal solution. The third-optimal assignment $A_3$ is generated and the cost is $C_3=8$, which is bigger than the cost of the current local optimal solution $\mathbb{P}_1$, which is 7. It indicates that it is impossible for $A_3$ to generate a better solution than  $\mathbb{P}_1$, and $\mathbb{P}_1$ is then confirmed as the global optimal solution. Till now, the optimal relative path set is got and the conflict-free motion sequences of vehicles in RCS is determined. Then, lower-level trajectory planning and tracking is conducted to switch the structure of the formation, and the process is given in Fig.~\ref{casee1}.

\subsection{Large-scale simulations}
\label{largescale}

In this part, the simulation is conducted and the results are analyzed. The settings of the simulation are firstly presented, and the reference rule-based method is designed. Then, box figure, heat maps, and snapshots are presented to show the results.

\subsubsection{Simulation settings}
\label{ss}

The simulation is implemented with MATLAB 2017b and SUMO 0.32.0~\citep{lopez2018microscopic} on a personal computer with CPU Intel CORE i7-8700@3.2GHz. The distribution and travel time of vehicles is analyzed to evaluate the performance. The parameters chosen for this simulation are presented in Table~\ref{para}. 

\begin{table}[htbp]
\centering
\caption{Simulation parameters}
\label{para}
\begin{tabular}{lll}
\toprule
Safe one-lane following gap                         &   $d_\text{F}$          & $15\,\mathrm{m}$ \\ 
Formation switching cycle                          &   $T_\text{F}$              & $4\,\mathrm{s}$ \\ 
Desired speed in formation                          &   $v_\text{F}$              & $15\,\mathrm{m/s}$\\ 
Minimum speed of vehicle                          &   $v_{\text{min}}$              & $0\,\mathrm{m/s}$ \\ 
Maximum speed of vehicle                          &   $v_{\text{max}}$              & $25\,\mathrm{m/s}$\\ 
Minimum acceleration of vehicle              &   $a_{\text{min}}$              & $-10\,\mathrm{m/s^2}$ \\ 
Maximum acceleration of vehicle                 &   $a_{\text{max}}$          & $5\,\mathrm{m/s^2}$ \\ 
Length of $S_1$                                                       &   $l_1$                           & $50\,\mathrm{m}$ \\ 
Length of $S_2$                                                       &   $l_2$                         & $350\,\mathrm{m}$ \\ 
Length of $S_3$                                                       &   $l_3$                          & $600\,\mathrm{m}$ \\ 
Parameters of rule-based method                      &   $d_\text{R}$          & $5\,\mathrm{m}$ \\ 
Parameters of rule-based method                      &   $\tau$                        & $0.66\,\mathrm{s}$ \\ 
Parameters of rule-based method                     &   $d_\text{end}$        & $150\,\mathrm{m}$ \\ 
\bottomrule  
\end{tabular}
\end{table}

The scenario chosen for the simulation is a three-lane road, and the three lanes are heterogenous and drivable for three different types of vehicles. The whole road consists of three segments: the initialization segment ($S_1$), the multi-lane driving segment ($S_2$), and the lane sorting segment ($S_3$). The length of the three segments is $l_1$, $l_2$, and $l_3$. Vehicles are generated at the start of $S_1$ randomly on the lanes with a constant speed $v_\text{F}$, and perform multi-lane driving in $S_2$ without considering lane preference. All lane changings are conducted in $S_3$ and must be finished before vehicles reach the end of $S_3$. Simulations are conducted in the three-lane scenario under different input traffic volume changing from $1000\,\mathrm{vehicle/(lane\cdot hour)}$ to $1600\,\mathrm{vehicle/(lane\cdot hour)}$ for the proposed and reference methods. For computational reasons, the number of vehicles in a single formation is limited to six, and relative motion mode 1 is chosen to regulate motion of vehicles, because more available motions can generate more conflicts and constraints in the searching tree, which significantly influences the computation time. The interlaced structure is chosen for formations to perform multi-lane driving and improve lane changing efficiency, and the parallel structure is chosen in $S_3$ after lane changing to improve capacity utilization.

\begin{figure}
\begin{center}
    \includegraphics[width=0.68\linewidth]{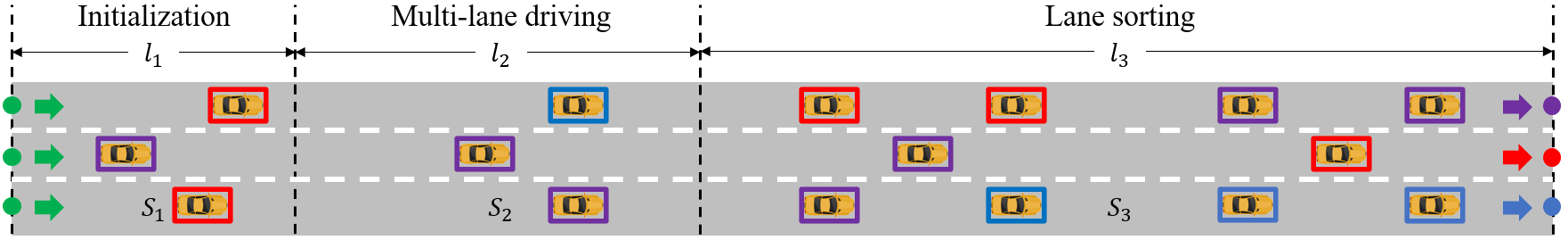}
    \label{simuscenario}
    \caption{The three-lane scenario of the simulation.}
    \label{cTree1}
\end{center}
\end{figure}

\subsubsection{Rule-based method}
\label{rbm}

We design a rule-based method as reference for performance comparison, which is also often considered in the existing research of lane sorting and intersection management. The designing of this rule-based method refers to existing lane-sorting and car-following methods~\citep{chouhan2020a,xu2021coordinated,bian2019reducing}, and guarantees no dead block at the end of the road which may happen in SUMO~\citep{sumoLaneChange}. Generally, vehicles which are not on their target lanes tend to slow down and try to find gap for lane changing as they approach the end of the road, and the slowing-down speed is set lower for vehicles that are closer to the end. The following rules are made for the rule-based method in this paper:

\begin{enumerate}[Rule 1.]
\item Vehicles drive with the constant speed $v_\text{F}$ and remain on their lanes if in $S_1$ or $S_2$, or has finished lane changing in $S_3$. Vehicles start to perform lane changing when reaching $S_3$ and have to finish lane changing before reaching the end of $S_3$.
\item The Constant Time Headway (CTH) policy is utilized as the minimum safe following gap between two adjacent vehicles in the same lane~\citep{bian2019reducing}. A vehicle performs lane changing when the gap between itself and the vehicles in the target lane satisfies CTH. The following gap $d_\text{R}$ calculated by the CTH policy is given as:
\begin{eqnarray}
d_\text{R}=d_0+v\times \tau,
\end{eqnarray}
where $d_0$ is the standstill gap, $v$ is the speed of the following vehicle, and $\tau$ is the time headway.
\item Vehicles which are closer to the end of $S_3$ are with higher priority than those which are farther. Vehicles with lower priority should slow down to let the vehicles with higher priority to perform lane changing at first. Noting the distance between a vehicle and the end of $S_3$ as $d_\text{end}$ and the distance that a vehicle must brake and stop to wait for others (in order to guarantee completeness of lane changing) as $d_\text{stop}$, the slow-down speed $v_\text{slow}$ of the vehicle with lower priority is determined as:
\begin{eqnarray}
v_\text{slow}=
\begin{cases}
v_\text{F}\times \frac{d_\text{end}-d_\text{stop}}{l_3-d_\text{stop}}, \ \text{if $d_\text{end}>d_\text{stop}$}, \\
0, \ \text{otherwise}.
\end{cases}
\end{eqnarray}
\end{enumerate}

\subsubsection{Results and discussion}
\label{rad}

\begin{figure}
\begin{center}
    \includegraphics[width=0.5\linewidth]{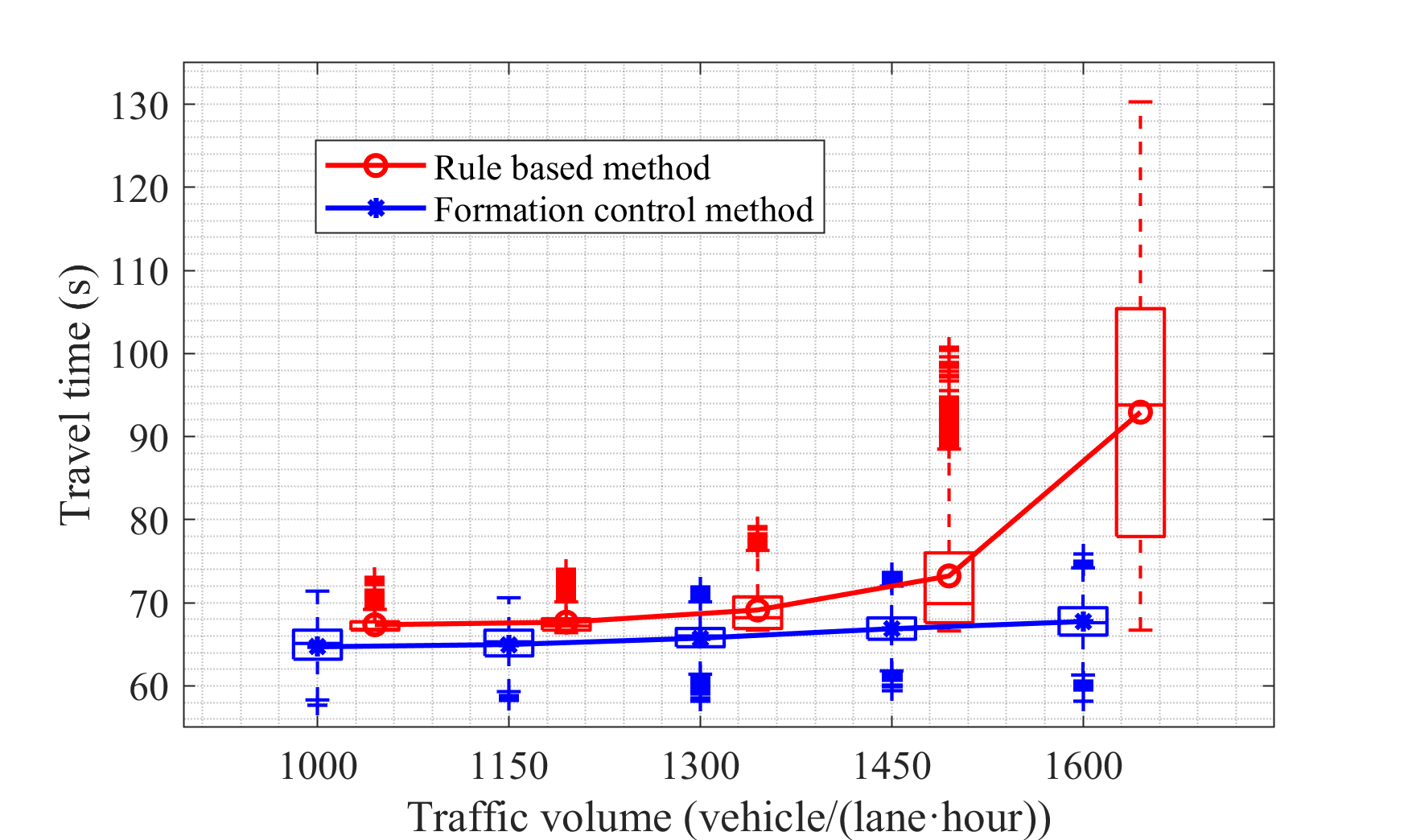}
    \caption{Results of travel time at different traffic volumes. The result of the rule-based method (red part) is shifted a little to the right to be distinguished from the result of the formation control method (blue part).}
    \label{boxfigure}
\end{center}
\end{figure}

The traveling time that vehicles pass the whole 1000-meter road is chosen to verify the performance of the proposed FC method and the rule-based method, as shown in Fig.~\ref{boxfigure}. The boxes represent the result distribution and the solid lines represent the average results. The red boxes represent results of the rule-based method, and the blue boxes represent that of the FC method. We didn't show the results at traffic volume lower than $1000~\mathrm{vehicle/(lane\cdot hour)}$ because the performance of the two methods is similar. The reason is that under low traffic volume, the distance between vehicles is long and few vehicles form formations. At higher traffic volume (not lower than $1000~\mathrm{vehicle/(lane\cdot hour)}$), the travel time of the rule-based method goes high, but the performance of the FC method keeps stable. The reason is that under these volumes vehicles may slow down for a long distance to meet a proper lane changing gap when using the rule-based method. We can also see that severe congestion happens under volume $1450~\mathrm{vehicle/(lane\cdot hour)}$ and $1600~\mathrm{vehicle/(lane\cdot hour)}$ when using the rule-based method. The results indicate that the proposed FC method can improve traffic efficiency and significantly reduce traffic congestion under high traffic volume, compared to the rule-based method.

\begin{figure}
\begin{center}
    \subfigure[$1150\,\mathrm{vehicle/(lane\cdot hour)}$]{
    \includegraphics[width=0.23\linewidth]{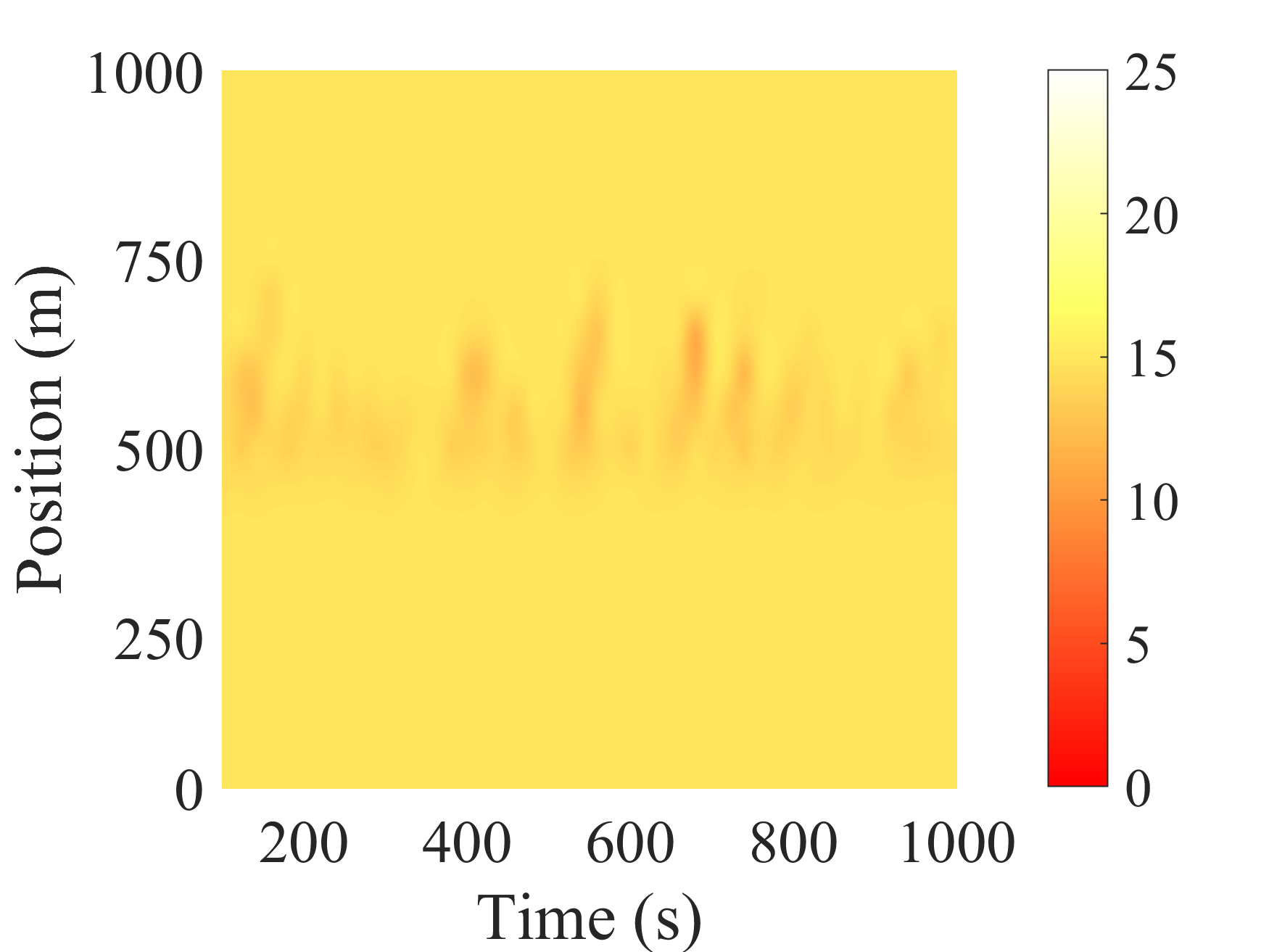}
    \label{ref100}}
\hspace{-2mm}
    \subfigure[$1300\,\mathrm{vehicle/(lane\cdot hour)}$]{
    \includegraphics[width=0.23\linewidth]{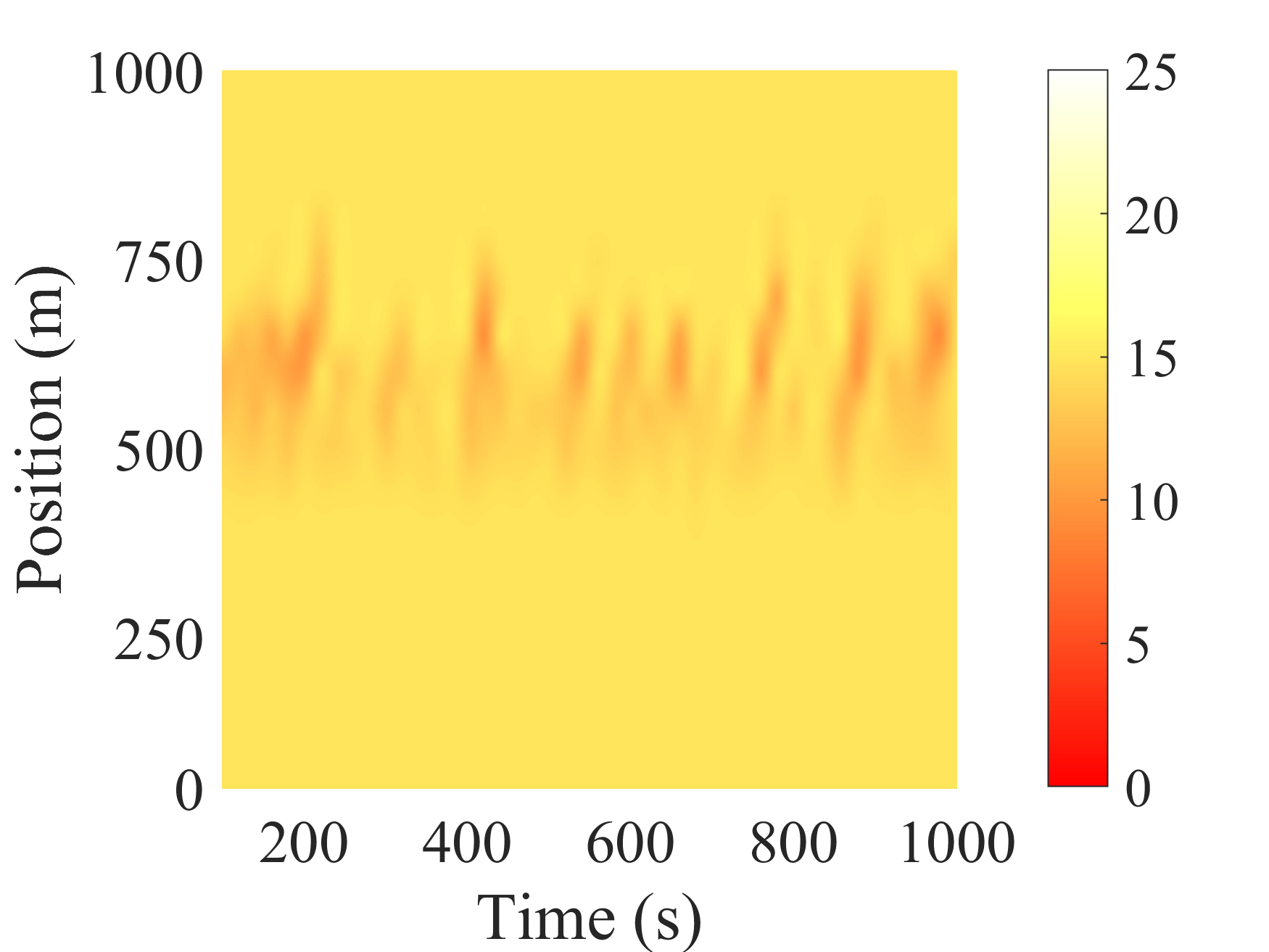}
    \label{ref112}}
\hspace{-2mm}
    \subfigure[$1450\,\mathrm{vehicle/(lane\cdot hour)}$]{
    \includegraphics[width=0.23\linewidth]{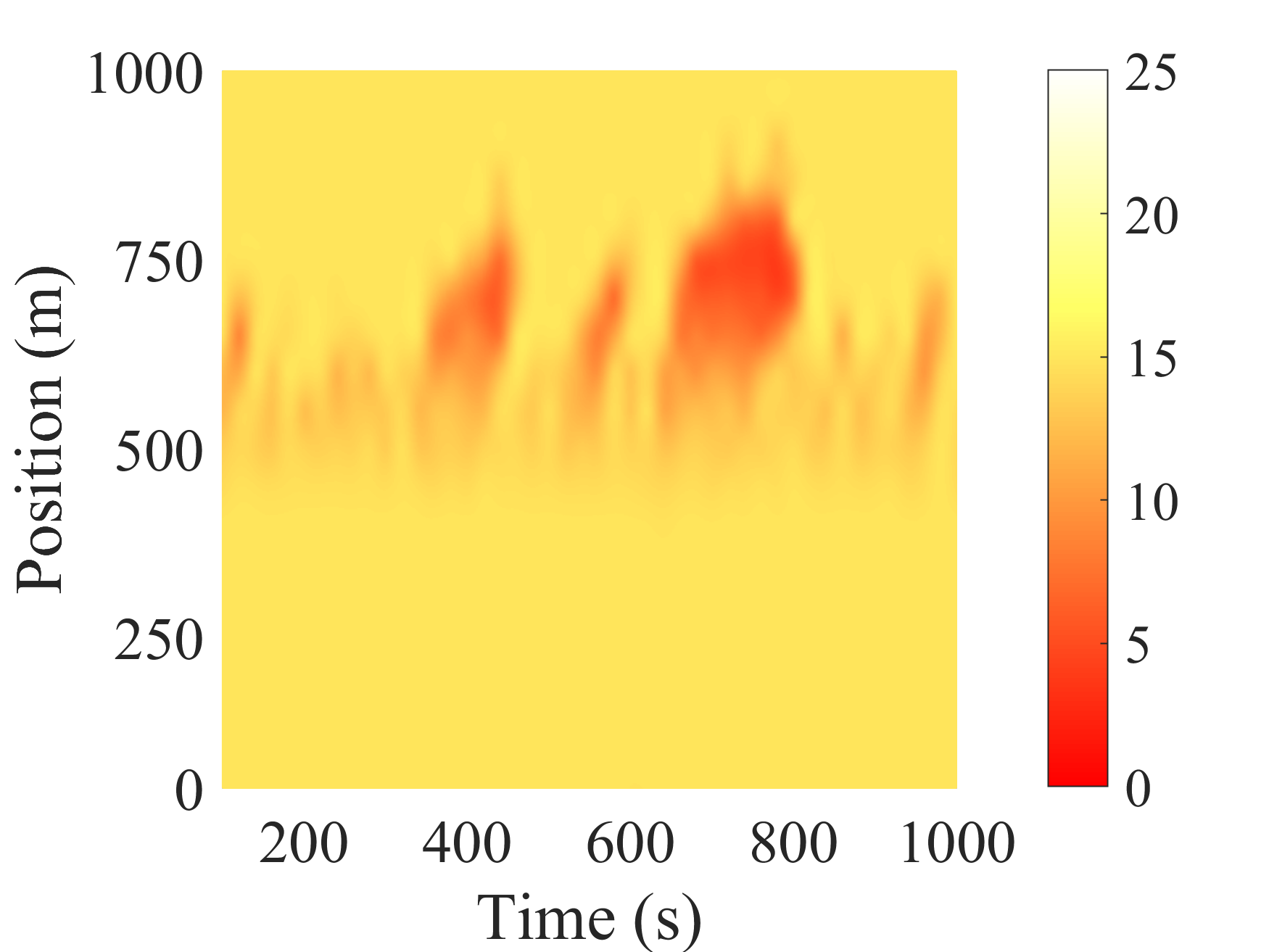}
    \label{ref125}}
\hspace{-2mm}
    \subfigure[$1600\,\mathrm{vehicle/(lane\cdot hour)}$]{
    \includegraphics[width=0.23\linewidth]{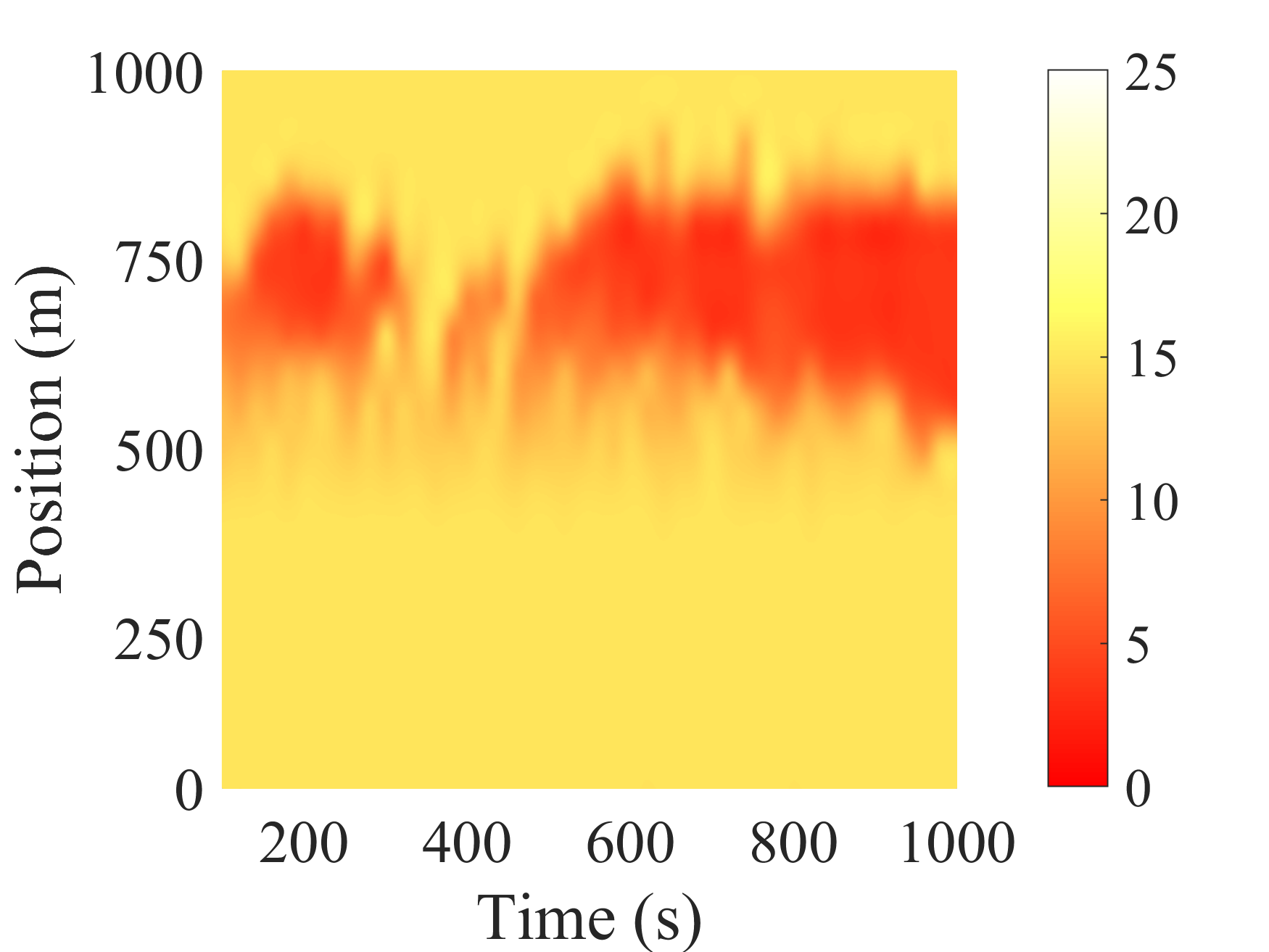}
    \label{ref137}}
    \subfigure[$1150\,\mathrm{vehicle/(lane\cdot hour)}$]{
    \includegraphics[width=0.23\linewidth]{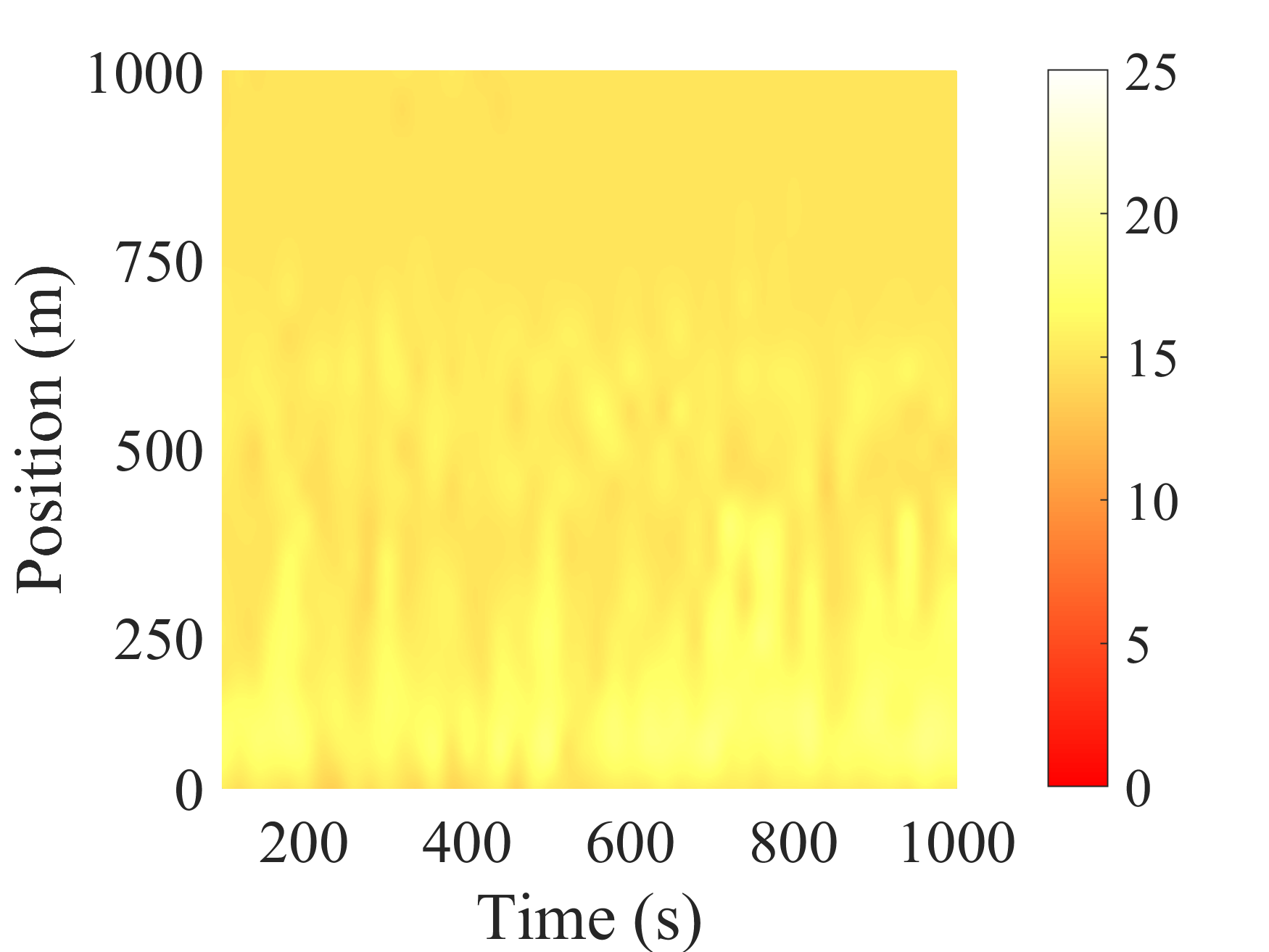}
    \label{frc100}}
\hspace{-2mm}
    \subfigure[$1300\,\mathrm{vehicle/(lane\cdot hour)}$]{
    \includegraphics[width=0.23\linewidth]{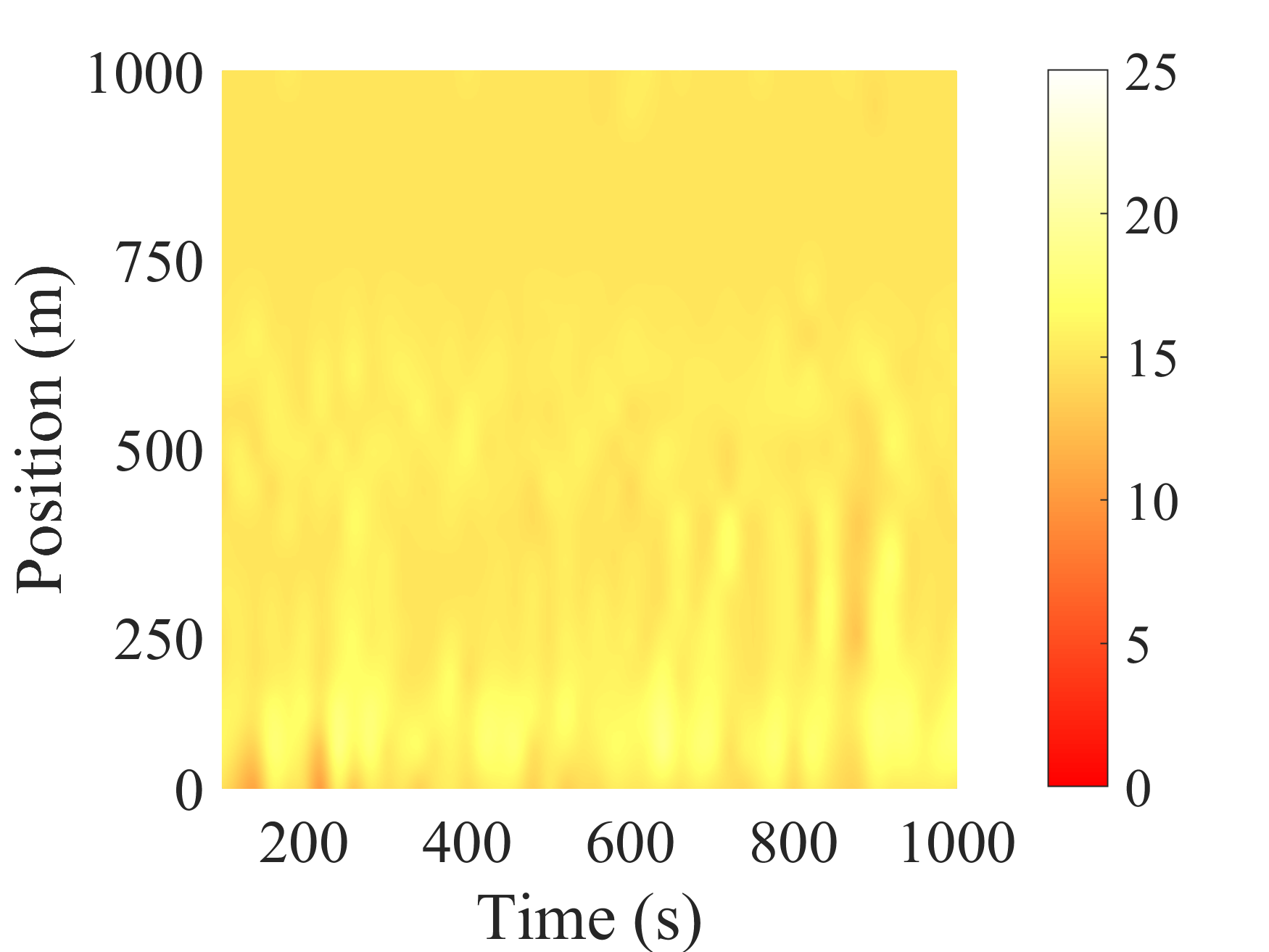}
    \label{frc112}}
\hspace{-2mm}
    \subfigure[$1450\,\mathrm{vehicle/(lane\cdot hour)}$]{
    \includegraphics[width=0.23\linewidth]{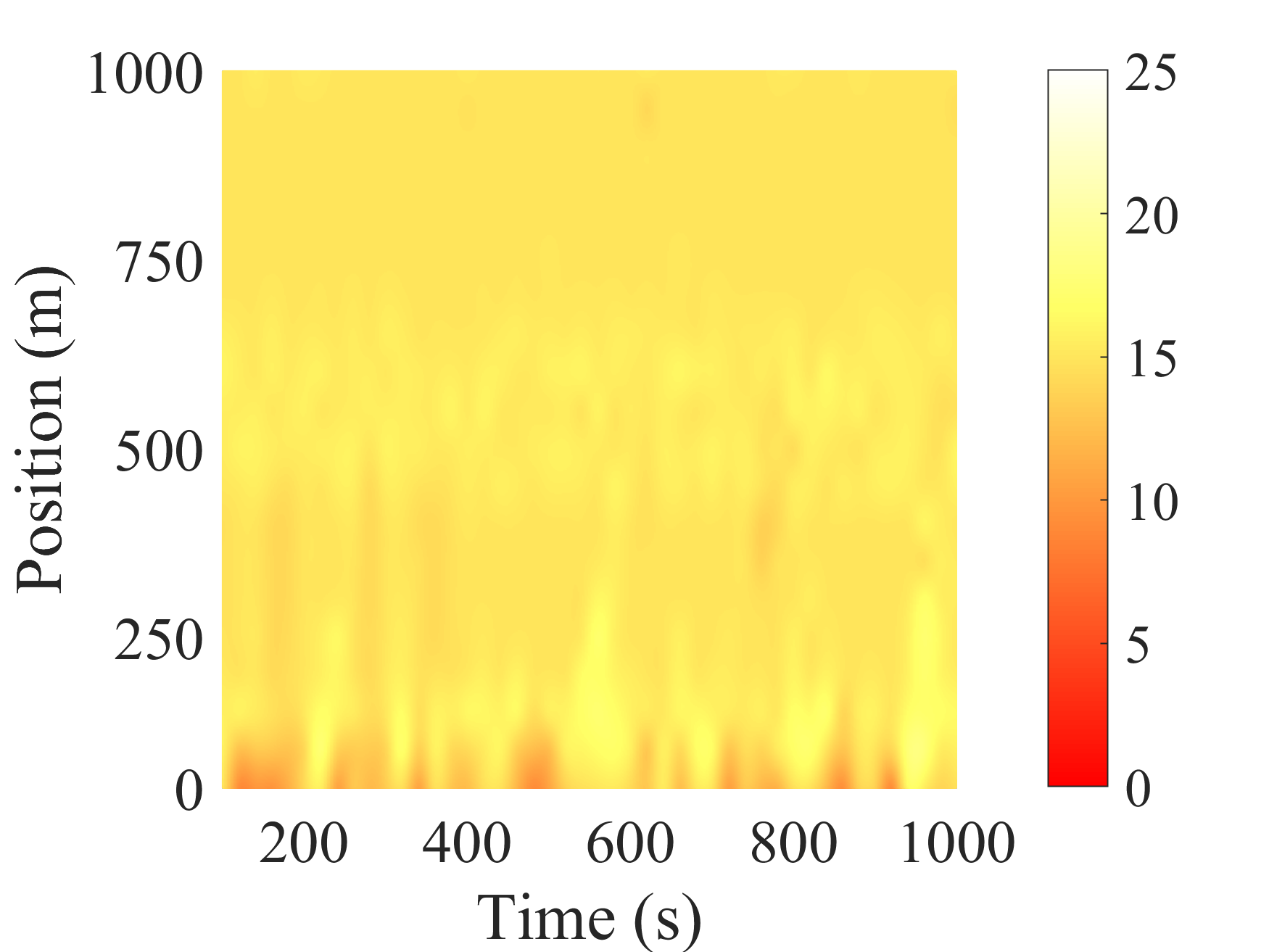}
    \label{frc125}}
\hspace{-2mm}
    \subfigure[$1600\,\mathrm{vehicle/(lane\cdot hour)}$]{
    \includegraphics[width=0.23\linewidth]{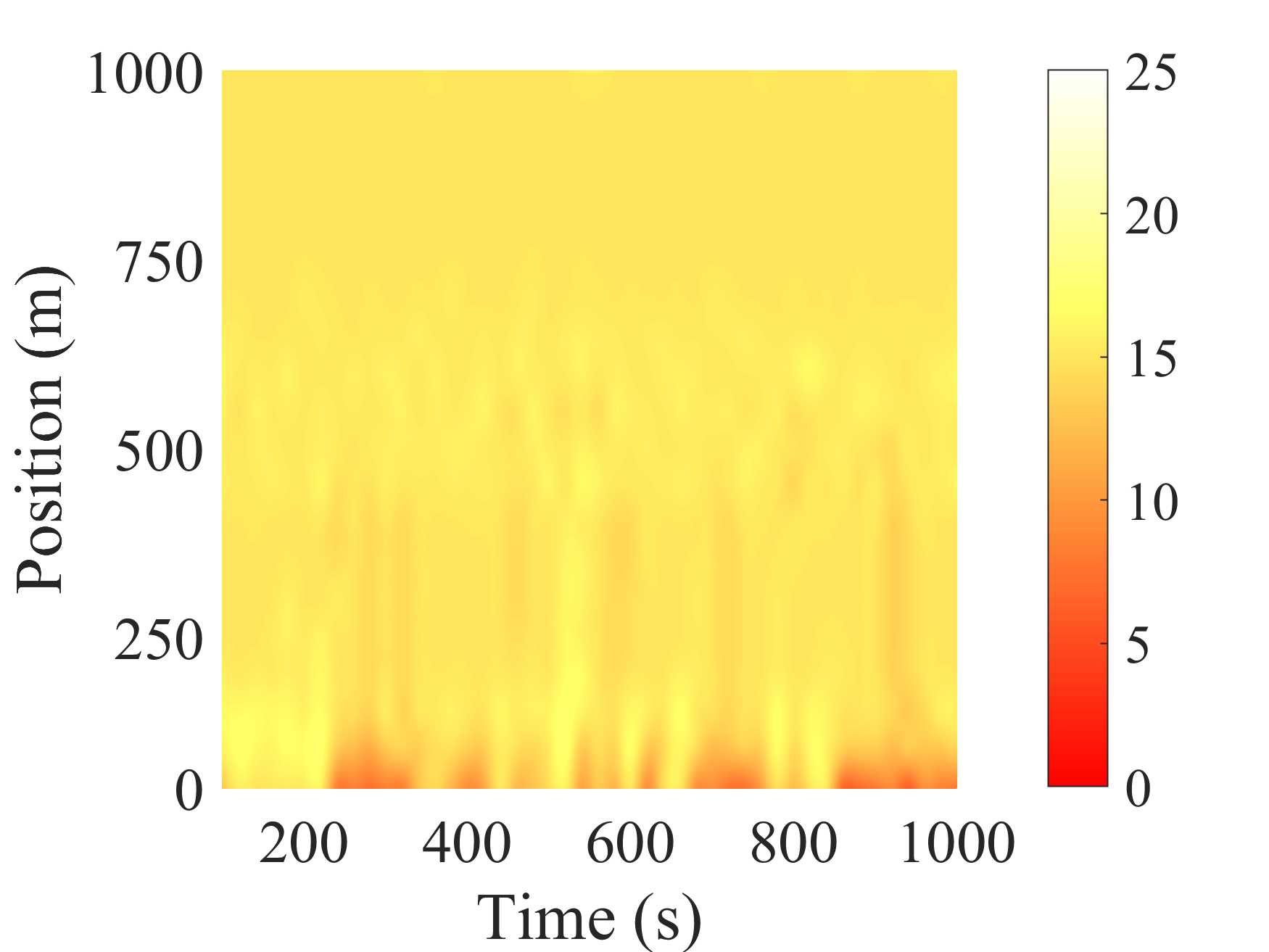}
    \label{frc137}}
    \caption{Heatmaps at different traffic volumes. The top four figures show the results of the rule-based method, and the bottom four figures show the results of the formation control method.}
    \label{heatmaps}
\end{center}
\end{figure}

\begin{figure}
\begin{center}
    \subfigure[Snapshots of the rule-based method]{
    \includegraphics[width=0.85\linewidth]{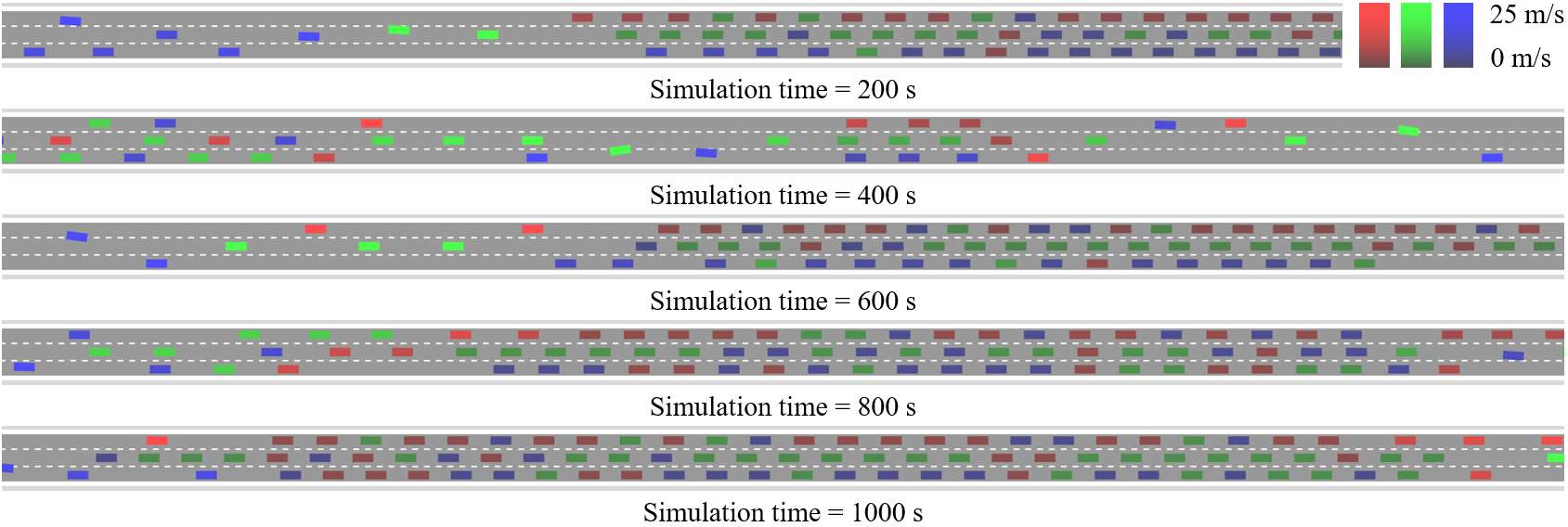}
    \label{ref}}
    \subfigure[Snapshots of the formation control method]{
    \includegraphics[width=0.85\linewidth]{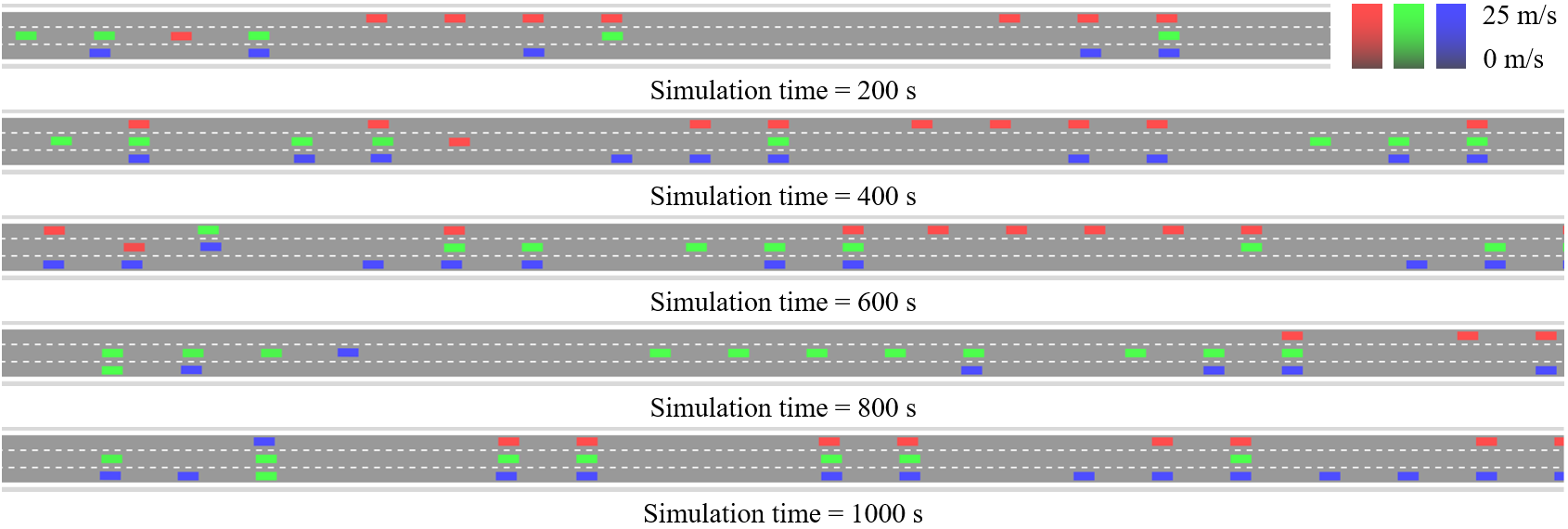}
    \label{frc}}
    \caption{Snapshots of the two methods in the area from $500~\mathrm{m}$ to $800~\mathrm{m}$ of the road at different simulation time. The full version video is available online at the address: {\color{blue}https://github.com/cmc623/Formation-control-with-lane-preference}}
    \label{snapshots}
\end{center}
\end{figure}

Heat maps are drawn to show the spatiotemporal speed distribution of vehicles, as shown in Fig.~\ref{heatmaps}. We show the heat maps at the four higher traffic volumes. The lighter color indicates that the average speed of vehicles in this spatiotemporal area is higher, and the darker indicates the slower. From the heatmaps we can see that congestion happens in the area from $500~\mathrm{m}$ to $800~\mathrm{m}$ of the road when using the rule-based method. The reason is that as vehicles approach the end of the road, the slowing-down speed to wait for lane changing gets lower, and when the traffic volume is high, the number of vehicles accumulates and congestion happens. However, when using the FC method, vehicles start to form formations when entering the road, and slight disturbance of speed may happen near the beginning area when traffic volume is high. After formations are formed, vehicles travel in groups with a constant desired formation speed and no severe congestion happens.

Snapshots are presented to show the distribution of vehicles when congestion happens, as shown in Fig.~\ref{snapshots}. The rectangles are vehicles and the color of vehicles represents their lane preference (red for the top lane, green for the middle lane, and blue for the bottom lane). The intensity of the color shows the speed of vehicles. The snapshots are taken under the highest simulation traffic volume ($1600~\mathrm{vehicle/(lane\cdot hour)}$) in the area from $500~\mathrm{m}$ to $800~\mathrm{m}$ of the road. From the snapshots we can see that severe congestion happens when using the rule-based method, which is consistent with the above conclusion of heat maps.

The theoretical upper bound of traffic volume $V_\text{u}$ of the proposed FC method can be calculated according to the formation speed and the following gap as:
\begin{eqnarray}
V_\text{u}=3600\frac{v_\text{F}}{2d_\text{F}}=1800\frac{v_\text{F}}{d_\text{F}}.
\end{eqnarray}

When the formation speed is $15~\mathrm{m/s}$ and the following gap is $15~\mathrm{m}$, the upper bound of the input traffic volume is $1800~\mathrm{vehicle/(lane\cdot hour)}$. When the input traffic volume is higher than the theoretical bound, congestion may happen before vehicles enter the 1000-meter scenario. However, we don't have to conduct simulation in that high volume because the proposed FC method already shows great improvement compared with the rule-based method at $1600~\mathrm{vehicle/(lane\cdot hour)}$.

%
\section{Conclusions}
\label{conc}
%

This paper proposes a formation control method that considers vehicles' preference on different lanes. The bi-level formation control framework is utilized to plan collision-free motion for vehicles, where relative target assignment and path planning are performed in the upper level, and trajectory planning and tracking are performed in the lower level. The collision-free multi-vehicle path planning problem considering lane preference is decoupled into two sub problems: calculating assignment list with non-decreasing cost and  planning collision-free paths according to given assignment result. The Conflict-based Searching (CBS) method is utilized to plan collision-free path for vehicles based on given assignment result. The case study and results of simulations indicate that the proposed FC method is able to plan collision-free path for vehicles considering their lane preference and significantly reduce congestion and improve efficiency at high traffic volume.

The  future  directions  of  this  research  include  extending the  proposed  FC  framework  to  ramp  merging  and  multi-lane intersection scenarios, and the stability analysis of the stable-driving formation with information or control disturbance.

%
\section*{Acknowledgment}
\label{ack}
This work is supported by the National Key Research and Development Program  of  China  under  Grant  2018YFE0204302,  the  National  Natural Science Foundation of China under Grant 52072212, and Intel Collaborative Research Institute Intelligent and Automated Connected Vehicles. 
%

\bibliographystyle{asmems4}
\bibliography{thesis}

\end{document}